\documentclass[a4paper,fleqn]{cas-dc}

\usepackage[numbers]{natbib}
\usepackage{svg}
\usepackage{amsmath,amssymb,amsfonts}
\usepackage{graphicx}
\usepackage{algorithm,algorithmic}
\usepackage{hyperref}
\usepackage{graphicx}
\usepackage{multirow}
\usepackage{placeins}
\usepackage{subcaption}
\usepackage{booktabs}
\usepackage{tabularx}
\usepackage{makecell}
\usepackage{fancyvrb}
\usepackage{xurl}
\usepackage{adjustbox}
\def\tsc#1{\csdef{#1}{\textsc{\lowercase{#1}}\xspace}}
\tsc{WGM}
\tsc{QE}

\begin{document}
\let\WriteBookmarks\relax
\def\floatpagepagefraction{1}
\def\textpagefraction{.001}

\shorttitle{SLIP}    

\shortauthors{B Podvin et~al.}  

\title [mode = title]{SLIP: Segmentation with Low-latency Interactive Prompting for 3D Medical Images}  



%


\author[1,2]{Baptiste Podvin} \cormark[1]
\cortext[1]{Corresponding author}
\ead{baptiste.podvin@ircad.fr}
\credit{Conceptualization, Methodology, Software, Data Curation, Investigation, Formal Analysis, Visualization, Writing – original draft}

\author[1]{Alexandre Ancel}
\credit{Methodology, Software}




\author[1, 3, 4]{Flavio Milana}
\credit{Data Curation, Resources}

\author[1,5]{Chiara Innocenzi}
\credit{Investigation}

\author[1, 5]{Davide Arrigo}
\credit{Investigation}

\author[1]{Federico Espinola Schulze}
\credit{Investigation}

\author[3, 4]{Guido Torzilli}
\credit{Data Curation, Resources}

\author[1]{Jacques Marescaux}
\credit{Funding Acquisition, Project Administration, Resources}

\author[2]{Daniel George}\fnmark[1]
\credit{Conceptualization, Project Administration, Resources, Supervision}

\author[1]{Alexandre Hostettler}\fnmark[1]
\credit{Funding Acquisition, Methodology, Project Administration, Resources}

\author[1]{Toby Collins}
\credit{Supervision, Methodology, Conceptualization, Project administration, Writing – review and editing}


\fntext[1]{These authors contributed equally.}


\affiliation[1]{organization={IRCAD France},
            city={Strasbourg},
            country={France}}


\affiliation[2]{organization={Université de Strasbourg, CNRS, ICube},
            city={Strasbourg},
            country={France}}

\affiliation[3]{organization={Division of Hepatobiliary and General Surgery, Department of Surgery, IRCCS Humanitas Research Hospital},
            city={Milan},
            country={Italy}}

\affiliation[4]{organization={Department of Biomedical Sciences, Humanitas University},
            city={Milan},
            country={Italy}}
\affiliation[5]{organization={Unità Operativa Complessa Ginecologia Oncologica, Dipartimento di Scienze Della Salute Della Donna e Del Bambino, Fondazione Policlinico Universitario A. Gemelli, Istituto di Ricovero e Cura a Carattere Scientifico (IRCCS)},
            city={Rome},
            country={Italy}}


\begin{abstract}
Interactive deep image segmentation enables efficient medical image annotation by iteratively refining predictions from user prompts, such as positive and negative clicks. Recent patch-based methods, including nnInteractive, achieve strong segmentation performance but remain limited in annotation workflows by high interaction latency, limited responsiveness to successive interactions, and the lack of support for reversible prompting. Furthermore, evaluation relies predominantly on simulated rather than controlled real-user interaction studies.

We present SLIP, an end-to-end trainable framework for interactive 3D medical image segmentation that decouples image encoding from prompt-guided refinement. Image features are computed once and reused, while a lightweight patch memory bank maintains an interaction-aware segmentation state shared across patches. This persistent representation enables prediction updates by propagating interaction context throughout the image, supports reversible prompting without recomputing image features, and substantially reduces interaction latency. By separating image representation from interactive reasoning, SLIP remains compatible with a wide range of image encoders.

We train a single SLIP model for general interactive segmentation across diverse anatomical structures and imaging modalities. Beyond standard simulated evaluation, we conduct a controlled prospective user study comparing manual segmentation, nnInteractive, and SLIP across three clinical annotation tasks, six expert participants, and subjective usability measures, addressing the limited human validation of interactive segmentation methods. SLIP achieves state-of-the-art interactive segmentation performance across 13 public datasets while providing lower interaction latency, greater responsiveness, support for reversible prompting, and higher user preference than existing approaches. Together, these findings demonstrate that annotation efficiency depends not only on segmentation accuracy but also on interaction design.

\end{abstract}




\maketitle

\section{Introduction, State-of-the-Art, and Contributions}

\label{sec:introduction}
\begin{figure}
    \centering
    \includegraphics[width=1.0\linewidth]{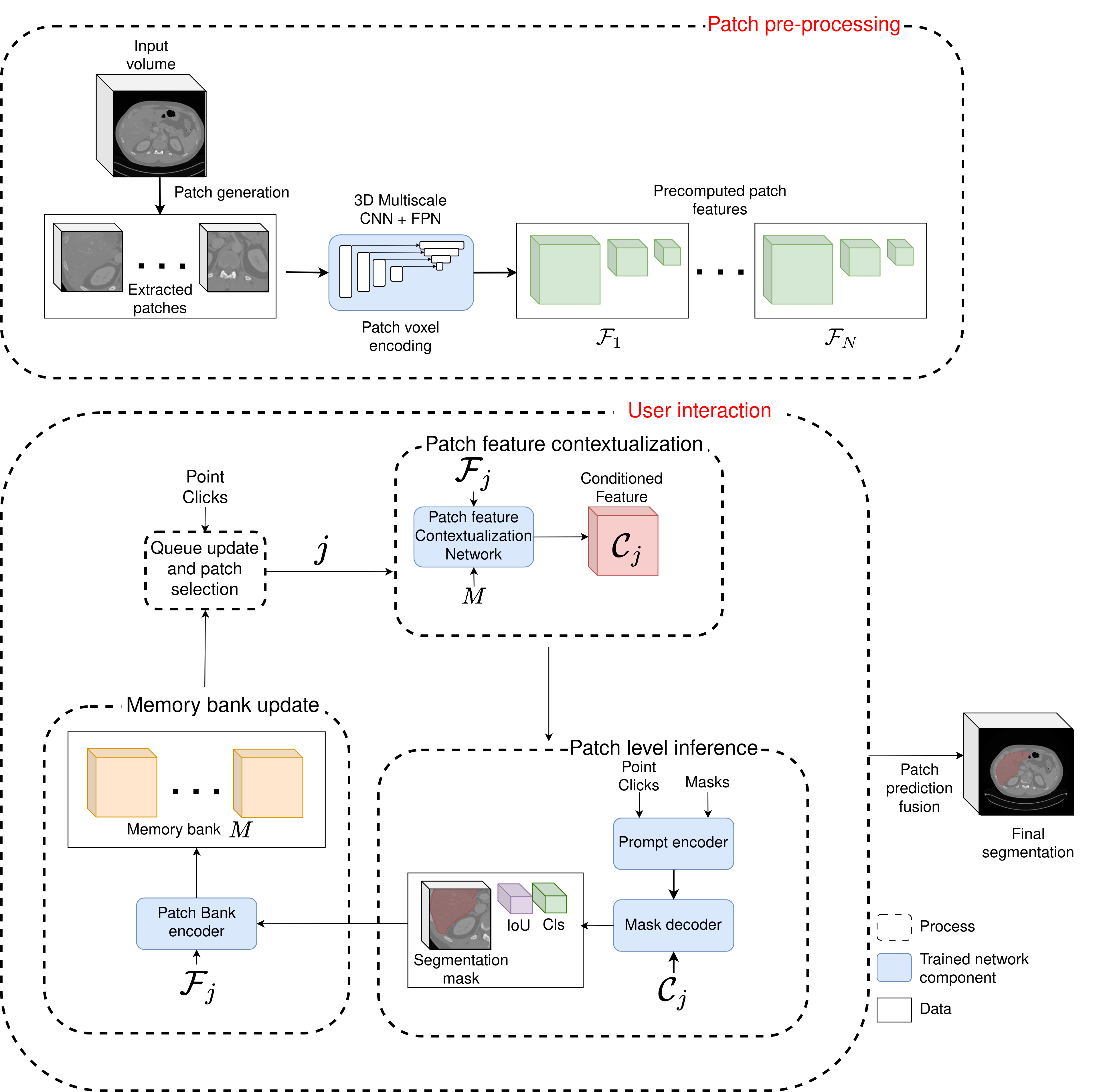}
    \caption{Overview of the SLIP inference pipeline and model architecture. Inference is divided into two stages. (1) Patch pre-processing: the input 3D image is partitioned into overlapping patches, each of which is encoded once by a shared CNN encoder to produce patch feature embeddings that are cached for the remainder of the interactive session. (2) Real-time user interaction: after each user prompt, only affected patches are revisited. A breadth-first patch queue selects target patches, whose cached features are contextualized using a lightweight Patch Contextualization Network conditioned on a fixed-capacity Patch Memory Bank containing information from previously inferred patches. A transformer-based mask decoder combines contextualized features with point prompts and the previous patch prediction to produce updated patch segmentations, uncertainty estimates, and patch-level classifications. The memory bank and patch queue are updated iteratively until no affected patches remain, after which overlapping patch predictions are fused into the final image-level segmentation.}
    \label{fig:architecture_overview}
\end{figure}

Promptable deep medical image segmentation methods use one or more human inputs (prompts), such as interior/exterior points, scribbles or bounding-boxes, to guide segmentation inference. In recent years, general-purpose segmentation models such as SAM~\cite{kirillov2023segment} for 2D images and its extension SAM-2~\cite{ravi2024sam} for video have significantly advanced interactive segmentation, enabling users to visualize and refine predictions in real time. These foundation models are trained on large-scale natural image and video datasets using synthetic prompts and ground-truth instance masks spanning thousands of object categories. However, despite their “segment anything” design, both SAM and SAM-2 exhibit limited generalization to medical imaging due to substantial domain shifts in appearance, modality, anatomy, and structural variability. Although fine-tuning on medical datasets can partially bridge this gap~\cite{cheng2023sam, zhu2024medical, ma2024segment, podvin2025samusa}, and SAM-2-style architectures have been extended to 3D medical segmentation by treating in-plane slices as video frames \cite{ma2025medsam2}, this formulation is suboptimal. In particular, the assumption of object permanence underlying SAM-2’s mask propagation is often violated across adjacent medical slices. Moreover, these models were not designed to accommodate changes in the imaging plane, limiting their flexibility in interactive 3D segmentation annotation workflows.

Recently, promptable segmentation methods have been proposed that operate directly on full 3D medical image volumes (3D-native), rather than processing data slice-by-slice, including SAMMed3D~\cite{wang2025sam}, SegVol~\cite{du2024segvol}, PRISM~\cite{li2024prism}, Vista3D~\cite{he2025vista3d}, and nnInteractive~\cite{isensee2025nninteractive}. To reduce computational cost and model complexity, these approaches typically partition the volumetric input into overlapping sub-blocks (patches), with inference performed at the patch level. Because each patch has a limited receptive field, mechanisms are required to propagate global context across patches. SegVol, for example, uses a single low-resolution patch as a global prior, though this can be insufficient for capturing fine structures such as vessels. In contrast, nnInteractive employs a dynamically sized, fixed-resolution region-of-interest (RoI) that is iteratively expanded as the predicted segmentation approaches the RoI boundary. This expansion is governed by a heuristic based on the number of foreground voxels contacting the RoI boundary. To recover high-resolution detail, nnInteractive further propagates RoI-level predictions into smaller overlapping patches at the original image resolution. In the recent Foundation Models for 3D Biomedical Image Segmentation (FM3BIS) CVPR 2025 challenge \cite{fm3bis2025}, which systematically benchmarks 19 methods including those above, nnInteractive achieved state-of-the-art performance among published methods in \textit{general interactive image segmentation}. This setting evaluates prompt-driven segmentation, where predictions are guided solely by user inputs rather than task-specific training, enabling segmentation of previously unseen structures beyond the training distribution.

Despite recent progress in general interactive image segmentation, state-of-the-art 3D-native methods, including nnInteractive (SOTA methods), exhibit several limitations:
\paragraph{Limitation 1: joint image–prompt encoding leading to inefficient interaction.}
SOTA methods rely on early joint encoding of image and prompt deep features within a tightly coupled pipeline. 

This coupling has two main drawbacks. First, voxel patches must be reprocessed through the full encoder decoder after each new prompt, increasing per-interaction computational cost and latency, particularly for large structures spanning multiple patches or under constrained hardware. Second, it limits reuse of pretrained image encoders, since image representations cannot be computed independently of prompts, reducing modularity and extensibility.

To address this, we propose to use decoupled image and prompt encoding, combined with a process we refer to as \emph{Patch feature contextualization}. Decoupled encoding enables image features to be precomputed and reused across interactions. It also promotes image encoder interoperability to exploit and test the latest foundation model image encoders. 
The idea of Patch feature contextualization is to iteratively refine frozen, precomputed patch features with information exchanged across neighboring patches during inference, through a learned contextualization function and a shared memory bank. Instead of re-encoding the image after each new prompt, each patch feature is updated at inference time using previously inferred segmentation and uncertainty signals, enabling prompt-driven information to propagate across patches while keeping the image encoder fixed and reusable.

\paragraph{Limitation 2: lack of reversible prompting.}

Current state-of-the-art interactive segmentation methods do not support reversible prompting, limiting users' ability to efficiently correct erroneous prompts or explore alternative segmentation strategies. As noted by the authors of nnInteractive, supporting prompt removal would require storing intermediate segmentation states because of architectural constraints, which becomes impractical for large volumes and resource-constrained devices~\cite{MICDKFZ_issue2_2025}. Consequently, interactions become effectively irreversible, reducing workflow flexibility and hindering iterative refinement.

To address this limitation, SLIP introduces reversible prompting through a native prompt-undo mechanism that allows users to retract and revise individual prompts without storing intermediate segmentation states. This enables efficient recovery from intermediate interaction errors while preserving previous annotation progress. In practice, we found reversible prompting to be particularly valuable for challenging segmentation tasks, where users progressively refine their understanding of the target anatomy through successive interactions and feedback from the model's intermediate predictions.

\paragraph{Limitation 3: limited real-user evaluation of interactive segmentation methods.}
SOTA method evaluation is largely based on simulated interaction. For example, the FM3BIS CVPR 2025 challenge uses virtual prompts that iteratively refine segmentations against ground truth, with performance measured after a fixed number of interactions. However, this setup does not capture real user behavior, including mistakes, self-corrections, ergonomic effects, or variability in interaction strategy. While nnInteractive includes one of the few real-user evaluations, it is limited to a single task (15 CT lesions), three clinicians, and does not control for case difficulty.

To address this, we conduct a controlled real-user study across three tasks, six users, and three methods (Vanilla Slicer, nnInteractive, SLIP), analyzed using a mixed-effects repeated-measures framework. We further include questionnaire based user experience measures capturing frustration and engagement.





\section{Methodology}
\subsection{Section organization}
The proposed architecture and inference methodology of SLIP are presented in Section~\ref{sec:methods_architecture_and_inference}. The end-to-end training methodology is presented in Section~\ref{sec:model_training}, and the training dataset is described in Section~\ref{sec:trainind_dataset}. In the main manuscript, we give detailed implementation of all novel network components, and unless otherwise specified, standard components, such as residual CNN backbones, follow established implementations. The supplementary material provides full implementation details for all components, and the code is publicly available at \url{https://github.com/IRCAD/SLIP}.

\subsection{Implementation Framework}
\label{sec:methods_impl_details}

All algorithms and models described in this work were implemented in Python 3.10 using PyTorch 2.2. To facilitate integration into existing clinical and research workflows, the graphical user interface (GUI) was developed as a segmentation plugin for 3D Slicer 5.6.2. The plugin extends Slicer's native segmentation capabilities, including image visualization, segmentation overlay and transparency control, and slice navigation with interactive point-click prompting, reversible prompting (implemented through prompt undo), and integrated backend inference.

To provide a streamlined user experience, patch-based processing is performed entirely in the background and is not exposed through the user interface. This abstraction allows users to focus on the interactive segmentation task while the underlying patch extraction, inference, and result aggregation are handled transparently by the system.

\subsection{SLIP inference process and model architecture}
\label{sec:methods_architecture_and_inference} 
The inference process is illustrated in Fig.\ref{fig:architecture_overview}, which operates in two consecutive phases: \textit{(i)} Patch pre-processing, then \textit{(ii)} Real-time User interaction. We first provide general notation definitions, then describe each step below.
\subsubsection{General notation and prompt semantics}
We denote the input 3D image as \(I(d,h,w)\), which may be scalar- or vector-valued. Without loss of generality, the first spatial dimension corresponds to the depth (slice) axis. The SLIP inference function is defined as $\hat{Y}_t = f(I,\mathcal{P}_t;\Theta)$, where $\Theta$ denotes the trainable parameters. The set $\mathcal{P}_t$ contains all user prompts available at time index $t$, ordered chronologically. 

The term $\hat{Y}_{t} \in [0,1]^{D \times H \times W}$ denotes the predicted binary segmentation map obtained from the input image $I$ and prompt set $\mathcal{P}_t$ via the model inference function $f$. SLIP’s inference function is composed of multiple sub-networks; we denote a sub-network by $f_{C}$, where $C$ is its identifier, with associated trainable parameters $\Theta_{C} \subset \Theta$.

At initialization ($t=0$), $\mathcal{P}_0=\emptyset$. Whenever a user provides a prompt, it is inserted at the head of $\mathcal{P}_t$ and the time index is incremented. Our model architecture can generalize to various prompt types, however, in this work we consider only binary-labeled point-click prompts (point-prompts), with a corresponding 3D position in image coordinates, and a user-supplied label. A foreground point (inside the target structure) has label 1, and a background point (outside the structure) has label 0. Since our objective is user-guided segmentation, prompt labels are treated as the source of truth that guides the predicted segmentation. Nevertheless, we support user-driven self-correction, allowing prompts in $\mathcal{P}_t$ to be flagged for removal. When this occurs, the prompt is removed from $\mathcal{P}_t$ and $t$ is incremented accordingly.

\subsubsection{Inference phase 1: Patch pre-processing}
\label{sec:image_preproc} 
\paragraph{Patch generation.}
Inference is performed by partitioning the image into \(N\) patches indexed by \(i \in [1, N]\), where \(I_i\) denotes the voxel cuboid corresponding to the \(i^{\text{th}}\) patch. Patches are extracted using a 3D sliding window of fixed size \(\mathbf{w} = (P_d, P_h, P_w)\). Image boundaries are handled via reflective padding.

Following nnU-Net, patches are generated with spatial overlap to enable smooth fusion of patch-level predictions into a final image-level prediction $\hat{Y}_{t}$. The overlap ratio along each dimension is defined as \(s \), expressed as a proportion of the patch size. We use \(s=90\%\), higher than  \(50\%\) used in nnU-Net, resulting in substantially lower patch density and therefore faster inference to support interactivity. Unlike nnU-Net, patches in SLIP are not processed in isolation but share contextual information across patches, reducing reliance on post-hoc smoothing of patch-wise predictions. Following SegVol, we adopt anisotropic patch sizes, with \(p=(32,192,192)\), reflecting the typical anisotropy of medical volumes, where the depth (slide) dimension is often substantially smaller than in-plane dimensions.

\paragraph{Patch voxel encoding.}
Image features are computed independently for each patch using a shared encoder \(f_E\): $\mathcal{E}_i = f_E(\tilde{I}_i; \Theta_E, \Phi_E)$, where \(\tilde{I}_i\) denotes the intensity-normalized patch. We apply z-score normalization, computing the mean and standard deviation over the full image \(I\) (i.e., image-level normalization). The encoder \(f_E\) is a standard architecture consisting of a 4-level multi-scale 3D residual CNN followed by a 4-level feature pyramid network (FPN)~\cite{lin2017feature}. As detailed in Section~\ref{sec:model_training}, the encoder parameters \(\Theta_E\) are learned end-to-end across multiple segmentation tasks, resulting in representations that capture general, structure-agnostic features useful for segmentation.

\subsubsection{Overview of Inference phase 2: Real-time User interaction}
\label{sec:overview_user_interaction}
The user interacts with SLIP through a graphical user interface (GUI), detailed in \ref{sec:methods_impl_details} , which enables visualization of the input image \(I\) and the current prediction \(\hat{Y}_t\) overlaid on \(I\). When the user adds or removes prompts, the prompt set  $\mathcal{P}_t$ is updated and passed to the inference pipeline, as illustrated in Fig.~\ref{fig:architecture_overview}. Inference proceeds iteratively by selecting a target patch, contextualizing its features, and predicting its segmentation. Concretely, each iteration consists of the following steps:

\paragraph{Target Patch selection.}
Patches are selected using a queue \(\mathcal{Q}\), where the patch \(j\) at the head is processed in the following steps. We initialize  \(\mathcal{Q}\) using all patches that spatially contain the most recent prompt in $\mathcal{P}_t$. To efficiently propagate context to neighboring patches, we update \(\mathcal{Q}\) in a breadth-first manner in the penultimate step. 

\paragraph{Patch feature contextualization.}
The patch features \(\mathcal{F}_j\) are transformed by the \emph{Patch feature Contextualization Network}, to produce new features \(\mathcal{C}^{t}_j\) that are adapted, or \emph{contextualized} to the specific segmentation task as communicated through the user prompts, and to predictions from other patches. The context information is stored in a fixed-size on-device memory bank \(M\). Initially, \(M=\emptyset\), and we update $M$ after the target patch has been processed, ready to contextualize the new head of \(\mathcal{Q}\).

\paragraph{Patch-level inference.}
The patch's contextualized features are input to a mask-refining decoder \(f_D\) with three heads that predict \textit{i)} the patch's segmentation map \(\hat{M}^{t}_j \in [0,1]^{P_d, P_h, P_w}\), segmentation uncertainty, and a binary patch-level classification (foreground \textit{vs.} background). \(f_D\) takes as input \(\mathcal{C}^{t}_j\), prompts in \(\mathcal{P}_t\) that fall within patch \(j\), and, when available, the patch's segmentation map prediction from the previous inference cycle, denoted as \(\hat{M}^{t-1}_j\), using prompt set $\mathcal{P}_{t-1}$.

\paragraph{Patch queue update.}
The queue \(\mathcal{Q}\) is updated by appending all patches that spatially overlap \(j\) (the one-hop neighbors of $j$). After predicting six negative patches, the queue is no longer updated. This enables efficient propagation of prompt and contextual information to neighbouring patches. 
\paragraph{Memory bank update}
The memory bank \(M\) is updated to support contextualization of the next patch in \(\mathcal{Q}\), using an allocation policy as described in Section \ref{sec:mem_bank_update}. 

The above steps are repeated until \(\mathcal{Q}\) is empty. Once completed, the patch-level predictions \(\{\hat{M}_1^{t}, \dots, \hat{M}_N^{t}\}\) are fused into the final image-level segmentation \(\hat{Y}_t\). We adopt the nnU-Net fusion strategy, where voxel-wise labels are obtained via Gaussian-weighted averaging of overlapping patch predictions, assigning higher weights to voxels near patch centers. The degree of smoothing is governed by the Gaussian standard deviation \(s_{\text{fuse}}\), set empirically to \(s_{\text{fuse}} = 0.125\).

The remainder of this section describes in detail Patch Feature Contextualization (Section~\ref{sec:patch_feature_contextualization}), Patch-Level Inference (Section~\ref{sec:patch_level_inference}), and Memory Bank Updating (Section~\ref{sec:mem_bank_update}).

\subsubsection{Patch feature contextualization}
\label{sec:patch_feature_contextualization}
We first describe the Patch Memory Bank, followed by the memory bank encoder, which produces a compact embedding of the memory bank. We then introduce the Feature Contextualization Network, which uses this embedding to contextualize the features of the currently selected patch. We refer to this patch as the \emph{target patch}, sampled from the head of \(\mathcal{Q}\).
The following formulation applies at every user interaction time; for clarity, we omit the time index. We denote by \(\mathcal{S}_{\text{infer}}\) the set of previously inferred patches. This is initially empty and updated after processing the target patch.

\paragraph{The Patch Memory Bank.}
The Patch Memory Bank \(M\) is a fixed-capacity on-device memory module that aggregates and transfers contextual information from \(\mathcal{S}_{\text{infer}}\). It captures complementary signals along three axes: (i) high-confidence predictions from strongly conditioned patches, associated with patches containing clicked prompts, (ii) temporal recency for propagating recently inferred states, and (iii) local spatial context induced by breadth-first traversal, where nearby inference steps tend to correspond to adjacent regions.

Accordingly, memory entries are organized into two groups:

\begin{itemize}
    \item \textbf{Group 1: High-confidence patches.} This group contains the highest-confidence patches in \(\mathcal{S}_{\text{infer}}\), ranked in descending order of confidence.
    \item \textbf{Group 2: Recently inferred patches.} This group contains the most recently processed patches, ordered by inference time. As described in Section~\ref{sec:overview_user_interaction}, breadth-first processing induces spatial locality, meaning that recently processed patches are often spatially close to the target patch.
\end{itemize}

In our implementation, the memory capacity is set to \(B = 6\) entries, with three entries per group. Each entry stores image features \(\mathcal{F}_k\), predicted mask \(\hat{M}_k\), an uncertainty estimate, and a binary patch-level classification \(\hat{v}_k\). For \(B = 6\), the total on-device footprint \(M\) is approximately 2 MB under Bfloat16 storage. Initially, \(M\) is empty and is updated after each patch inference step, as described in Section~\ref{sec:mem_bank_update}.

\paragraph{Patch Memory Bank Encoder.}

We encode each memory bank entry independently with a shared lightweight CNN encoder. First, \(\hat{M}_k\) is downsampled by three strided convolutions, each followed by spatial LayerNorm and GELU, then projected with a final convolution to \(d_m\), yielding \(\mathcal{M}_k\); this is fused with \(\mathcal{F}_k\) by element-wise addition and channel-aligned with a convolution.

The fused features are refined using \(C\) ConvNeXt-style blocks \cite{woo2023convnext}, each consisting of LayerNorm normalization, a depthwise convolution with kernel size \(k=7\), and GELU activation.

To incorporate metadata, an object-presence embedding \(e_k^{(v)}\) is derived from the binary patch-level classification \(\hat{v}_k\) using a learned embedding. This embedding is appended as an additional token per entry, forming an augmented token sequence.

Finally, entry-wise positional \(p_k\) are injected via addition, where \(p_k\) is parameterized using learned embeddings, encoding temporal order. The high-confidence patches are encoded with a fixed position.

\paragraph{Patch Contextualization Network.}
We implement \(f_C\) as a stack of \(L_C\) \textbf{Patch Attention Layers}. Each layer refines the contextualized features via self-attention over patch features \(\mathcal{F}_j\), and cross-attention over a single memory bank token $\mathbf{t}_M$; a variable-length feature vector formed by concatenating and flattening the encoded memory bank entries. 

Each Patch Attention Layer consists of the following sequential blocks:

\begin{itemize}
    \item \textbf{Self-attention block.}  
    This performs full self-attention on \(\mathcal{F}_j\), using a single attention head with embedding dimension \(256\).
    \item \textbf{Cross-attention block.}
    This performs cross-attention, where \(\mathcal{F}_j\) attends to \(\mathbf{t}_M\). A single attention head is used, and to decrease computational cost and memory usage, the key and value representations are projected to a reduced dimensional space of size \(64\).
    \item \textbf{MLP block.}  
    This produces the final contextualized features, implemented as a two-layer feed-forward network with hidden dimension \(2048\) and ReLU activation.
\end{itemize}

%

%
The output of the final Patch Attention Layer produces \(\mathcal{C}\), which is then used for patch-level inference. 

\subsubsection{Patch-level inference using the mask decoder}
\label{sec:patch_level_inference}
The mask decoder predicts and refines the segmentation \(\hat{M}_k\) of the target patch using three complementary information sources: (1) \textbf{Target-specific point-prompts}, using point-prompts $\mathcal{P}_k \subset \mathcal{P}$ within the target patch, (2) (when available), the segmentation estimate \(\hat{M}'_k\) from the previous inference of the target patch, and (3) the contextualized patch features \(\mathcal{C}_k\). We express (1) as a variable-length token stream, and (2) as a segmentation prompt. The decoder outputs three prediction tokens: the segmentation mask \(\hat{M}_k\), an IoU prediction token \(\hat{u}_k\), and a classification token \(\hat{c}_k\).

\begin{equation}
\begin{split}
(\hat{M}_k, \hat{u}_k, \hat{c}_k)
= f_D(&\mathcal{E}_{P\text{-}Prompt}, \mathcal{E}_{M\text{-}Prompt}, \\
&\mathcal{C}_k; \Theta_D)
\end{split}
\end{equation}

\noindent The token set $\mathcal{E}_{PointPrompt} = \{e^1_{\text{prompt}}, \dots, e^N_{\text{prompt}}\}$ denotes the \emph{point prompt tokens}, each of which is a real-valued vector of size $d_{point}$, sorted by prompt recency. If $\mathcal{P}_k$ is empty, we place in $\mathcal{E}_{PointPrompt}$ a learnable token (the ``no-click'' token) of size $\ensuremath{d_{point}}$. This provides a valid prompt representation for all target patches.

\noindent The token set $\mathcal{E}_{MaskPrompt}$ represents \(\hat{M}'_k\) as a single segmentation prompt token of dimension $\ensuremath{d_{mask}}$. If \(\hat{M}'_k\) is not yet available, we use a single learnable token (the ``no-mask'' token) of dimension $\ensuremath{d_{mask}}$.

\noindent The output token \(\hat{u}_k\) corresponds to a learned IoU prediction token that estimates the intersection-over-union quality of the predicted segmentation \(\hat{M}_k\). The output token \(\hat{c}_k\) corresponds to a learned classification token used to predict the object presence in the patch.

\paragraph{Tokenization implementation.}

If $\mathcal{P}_k \neq \varnothing$, each prompt is tokenized from its 3D location $\mathbf{p}\in[-1,1]^3$ in normalized patch coordinates.  
For each prompt, we extract a local visual descriptor $\mathcal{D}$ by trilinear point sampling of patch features $\mathcal{F}_k$ at the centered coordinate $\mathbf{p}_c=\mathbf{p}+0.5$, and combine it with positional encoding to form the prompt token.  
This visual sampling complements coordinate cues with local appearance, yielding tokens that are both spatially grounded and semantically discriminative, which is particularly important in 3D where coordinate-only embeddings are more susceptible to attention-level information dilution.


We then encode $\mathcal{D}$ as follows:
\begin{equation}
e_{\text{prompt}} = e_{\text{pos}}(\mathbf{p}_c) + \phi_A(\mathcal{D}) + e_{\text{type}}(l),
\end{equation}
\noindent where $e_{\text{pos}}(\mathbf{p}_c)$ is a random Fourier positional encoding of the normalized 3D coordinate, $\phi_A$ is the trilinearly sampled visual feature vector at $\mathbf{p}_c$, and $e_{\text{type}}(l)$ is a learned label embedding (positive, negative, or not-a-point).

If $\hat{M}'_k$ is available, its logits are encoded by two strided Conv3D\(+\)LayerNorm3d\(+\)GELU blocks, followed by a final \(1\times1\times1\) Conv3D projection to \(d_e\), yielding dense mask tokens at image-embedding resolution.


\paragraph{Mask decoder architecture.}
The mask decoder architecture is strongly inspired by SAM-based 2D models, with explicit adaptations to 3D data: \(f_D\) is a 3D two-way transformer decoder (depth \(=2\), \(8\) heads) that consumes dense image tokens (additively fused with dense mask tokens) and prompt tokens (concatenated with output tokens: object-presence, IoU, and mask tokens), then produces three outputs through dedicated heads, namely segmentation prediction, segmentation confidence (IoU quality), and patch-level binary classification (object presence); during inference, updated mask tokens are mapped by lightweight MLPs and applied to an upscaled dense 3D feature map (two transposed-convolution stages) to generate volumetric mask logits.

\subsubsection{Patch memory bank update}
\label{sec:mem_bank_update}
After inferring the target patch, if $\mathcal{Q}$ is not empty, the memory bank is updated for the next target at the head of $\mathcal{Q}$. Recall that the memory bank uses two patch groups: Group 1: high-confidence patches, and Group 2: recent patches (Section \ref{sec:patch_feature_contextualization}). Each group is updated as follows:

\paragraph{Group 1: High-confidence patches.}
First, we filter out patches that do not contain (spatially intersect) at least one prompt in $\mathcal{P}_t$. Of these, low-confidence patches are filtered out via thresholding: $\hat{u}^t_i < \tau_u \in \Psi$ and $\hat{v}^t_i < \tau_v \in \Psi$, where $\hat{u}^t_i$ and $\hat{v}^t_i$ are the patch's predicted IoU and mask-stability score. The confidence thresholds are set empirically to $\tau_u = 0.5$ and $\tau_v = 0.8$. Then up to three patches are allocated to the memory bank, sorted in confidence order using a combined confidence score $c_i = w_u \, \hat{u}^t_i + \hat{v}^t_i$ where $w_u \in \Psi = 1.5$.
\paragraph{Group 2: Recently inferred patches}
A first-in-first-out policy is used, where the most recently inferred patch is moved into the memory bank, displacing the patch that was least recently inferred. Patches belonging to Group~1 are not inferred and are therefore not appended to the memory bank.

\subsubsection{Reversible prompting}
\label{sec:prompt_undo}

SLIP supports \emph{reversible prompting}, allowing users to sequentially remove previously issued prompts with no limit on the number of undo steps. This enables users to efficiently correct interaction mistakes, revise segmentation strategies, and explore alternative refinement trajectories without restarting the annotation process.

A naïve implementation would reconstruct a previous interaction state by replaying the complete sequence of user prompts, resulting in computational cost that increases linearly with the number of interactions. Instead, SLIP caches the prediction state of each affected patch after every interaction step $t$, enabling exact and immediate restoration of previous interaction states.

This approach is practical because of SLIP's lightweight architecture, particularly the separation of image and prompt feature encoding, which substantially reduces the memory required to cache intermediate states. In contrast, caching equivalent intermediate states for architectures such as nnInteractive would be prohibitively expensive; for example, storing the intermediate states for a typical CT segmentation containing 20 interactions would require approximately 20~GB of memory.

Memory requirements are further reduced by exploiting the sparsity of user interactions. Because each prompt affects only the subset of patches contained in the active patch queue $Q$, SLIP stores intermediate states only for these affected patches together with their corresponding mask decoder outputs. Furthermore, patch updates are stored as compressed deltas rather than complete prediction states. Consequently, storing the complete interaction history for a typical CT segmentation with 20 interactions requires approximately 140~MB of memory.

Following restoration of a previous interaction state, patch fusion is performed and the updated segmentation is immediately displayed to the user through the graphical interface.

\subsection{Model Training}
\label{sec:model_training}

SLIP was trained from scratch using a multimodal dataset of 3D medical images and corresponding annotated segmentation label maps. Foreground patches are randomly sampled from regions with ground-truth annotations. Following prior work, label 0 was reserved for the background class, while non-zero integer labels represented foreground anatomical structures. Training was performed on an NVIDIA DGX A100 server equipped with four GPUs, each providing 40~GB of VRAM.

Training was performed using supervised learning with the mask decoder loss defined below. A two-stage training strategy was adopted. Stage~1 trained prompt-conditioned patch-wise segmentation with the memory bank disabled, whereas Stage~2 additionally trained memory-based semantic propagation across neighboring patches.

\subsubsection{Mask Decoder Loss}

The mask decoder loss supervised the decoder's three prediction heads: mask prediction, mask confidence prediction, and patch-level classification. The individual loss terms were defined as follows:

\begin{itemize}
    \item \textbf{Mask prediction loss}: Computed as a weighted combination of focal ($\mathcal{L}_{\mathrm{Focal}}$) and Dice Similarity Coefficient (DSC) ($\mathcal{L}_{\mathrm{DSC}}$) losses.
    \item \textbf{Mask confidence loss} ($\mathcal{L}_{\mathrm{Conf}}$): Computed as the mean absolute error (MAE) between the predicted and ground-truth IoU.
    \item \textbf{Classification loss} ($\mathcal{L}_{\mathrm{BCE}}$): Computed using binary cross-entropy.
\end{itemize}

The overall mask decoder loss was defined as

\begin{equation}
\mathcal{L}_{MD}
=
\lambda_{\mathrm{Focal}}\mathcal{L}_{\mathrm{Focal}}
+
\lambda_{\mathrm{DSC}}\mathcal{L}_{\mathrm{DSC}}
+
\lambda_{\mathrm{Conf}}\mathcal{L}_{\mathrm{Conf}}
+
\lambda_{\mathrm{BCE}}\mathcal{L}_{\mathrm{BCE}}
\end{equation}

where $\lambda_X$ denotes the corresponding loss weight. DSC and focal losses were combined to exploit their complementary strengths: DSC loss is robust to severe class imbalance, whereas focal loss emphasizes difficult examples and hard-to-segment regions. Because the focal loss typically has a smaller numerical magnitude than the DSC loss, its weight was increased to 20.0 so that both terms contributed comparably during optimization while still emphasizing difficult voxels. All remaining loss weights were set to 1.0.

For patches containing no foreground voxels, only the classification loss was optimized, while all segmentation-related loss terms were set to zero. This avoided ill-conditioned DSC- and IoU-based supervision on empty targets while still training the model to distinguish foreground from background patches.

\subsubsection{Stage 1: Patch-wise Training}

During Stage~1, the memory bank was reset before processing each patch and remained empty throughout the forward pass, preventing information propagation between patches. Consequently, prompt-conditioned segmentation was supervised using only information from the current patch and the simulated prompts. Simulated prompts were used to mimic iterative corrective user interactions within each patch.

The initial prompt set, $\mathcal{P}_i^1$, consisted of a single point sampled uniformly from all foreground voxels in patch $i$. The label at the sampled voxel determined the foreground anatomical structure, while all remaining labels were treated as background, thereby defining the target binary segmentation mask.

A forward pass of the user interaction procedure was performed on patch $i$ using the initial prompt set $\mathcal{P}_i^1$. The resulting prediction was supervised using $\mathcal{L}_{MD}$.

Subsequent prompt sets were generated iteratively by adding a single corrective point to the previous prompt set. Using the current mask prediction, the dominant error type was determined by comparing the numbers of false-positive and false-negative voxels. If false negatives dominated (under-segmentation), the new prompt was assigned a positive label; otherwise, it was assigned a negative label. The prompt location was sampled uniformly from voxels belonging to the dominant error region.

The updated prompt set was then used in a subsequent forward pass on patch $i$, and the resulting loss was accumulated. This procedure was repeated until a maximum of 16 prompts had been simulated, after which the accumulated loss was backpropagated. This upper limit provided sufficient simulated corrective interactions while maintaining practical training times.

Patch-wise training used a batch size of $B=32$ patches. Class imbalance was handled via dataset-level class weights, while patches were randomly sampled from annotated structures without explicit class balancing during sampling. Prompt simulation and loss accumulation were performed independently for each patch before the accumulated batch loss was backpropagated.

Model parameters were optimized using AdamW with an initial learning rate of $4\times10^{-5}$. A cosine annealing learning-rate schedule was applied throughout training. We trained for 300 epochs using bfloat16 mixed precision and $\ell_2$ gradient clipping with a maximum norm of 0.1 across four GPUs.

\subsubsection{Stage 2: Full-model Training}

Stage~2 retained the prompt-conditioned supervision used during Stage~1 while additionally supervising neighbouring patches using semantic information propagated through the memory bank. Consequently, the same loss was applied in two complementary settings: to prompted patches with the memory bank reset, and to unprompted neighbouring patches conditioned on the evolving memory bank.

A straightforward end-to-end implementation of the interactive inference procedure was not computationally feasible during training. Simulating user interactions while dynamically growing the patch queue would require retaining the entire computation graph across successive queue insertion operations, resulting in prohibitively high GPU memory consumption. Instead, fixed-length patch queues were sampled directly from the training volumes, with patch insertion disabled. Consequently, gradients were propagated only through the patches explicitly included in each sampled queue.

Two batch types were sampled with equal probability throughout training:

\begin{itemize}
    \item \textbf{Context-sharing batches}, which supervised both prompted head patches and memory-conditioned neighbouring patches.
    \item \textbf{Non-context-sharing batches}, which supervised only prompted patches with the memory bank reset, identical to Stage~1.
\end{itemize}

Patches with fewer than seven spatial neighbours, typically originating from lower-resolution images, were eligible only for non-context-sharing batches.

\paragraph{Memory-conditioned supervision}

Each context-sharing batch comprised $B_q =8$ queues, each containing up to eight spatially neighboring patches. A queue was constructed by randomly sampling a patch containing at least one foreground voxel as the head of the queue and filling the remaining positions with its neighboring patches. Only the head patch underwent prompt simulation, avoiding redundant prompt supervision of patches that may appear in multiple sampled queues.

The head patch first underwent the same corrective prompt simulation procedure used during Stage~1. After the final simulated interaction, its resulting memory representation was used to initialize the memory bank.

The remaining neighboring patches were then processed sequentially without prompts. For each neighboring patch, the model predicted the segmentation using the current patch together with semantic information retrieved from the memory bank. The loss was evaluated against the ground-truth segmentation, after which the memory bank was updated before processing the next patch. This procedure was repeated until all neighboring patches in the queue had been processed, progressively propagating semantic information throughout the sampled neighborhood.

During training, one modification was made to the inference data flow: the memory bank was updated using the final patch-wise prediction rather than the mask decoder output. This produced a more stable representation for context sharing because starting from an under-refined segmentation would cause the memory bank to encode ambiguous object boundaries, which cross-attention would then replicate across neighboring patches, compounding spatial errors rather than correcting them. By ensuring the seed patch reaches a high-confidence segmentation before contextualization begins, the memory bank reliably encodes a clean object representation that the contextualization mechanism can learn.

\paragraph{Training details}

Training on the DGX server used $B_q = 8$ queues per batch and a queue length of eight patches, requiring 144~GB of GPU memory (36~GB per GPU). A total of 500 queues per epoch were sampled. The final checkpoint obtained from Stage~1 was used to initialize Stage~2.

\subsubsection{Data Augmentation}

The same image-level augmentation pipeline was used during both Stage~1 and Stage~2 training. The augmentation strategies are summarized below, while complete implementation details and parameter ranges are provided in the supplementary material (Table 4).

\textbf{Intensity augmentation.} Random intensity transformations included brightness and contrast adjustment, gamma correction, additive Gaussian noise, and Gaussian blurring. For MRI images, random bias field simulation, ghosting artefacts, and k-space spike artefacts were additionally applied.

\textbf{Geometric augmentation.} Random geometric transformations included resampling, resizing, flipping, rotations, and elastic deformations.

To improve the diversity and robustness of the training labels, two label augmentation techniques were employed.

\textbf{Pseudo-labelling.} Pseudo-labels were introduced following the strategy proposed by nnInteractive. A 2D image slice was first partitioned into supervoxels using SAM auto mask generation, after which one supervoxel was randomly selected as the pseudo-foreground region. This pseudo-label was subsequently propagated to neighbouring slices using SAM2, thereby generating additional weakly annotated training examples.

\textbf{Label hierarchy randomization.} Point-based prompts may be ambiguous when anatomical structures are organized hierarchically (e.g., vessels within organs or lesions within surrounding tissue). To improve robustness to such ambiguity, hierarchical labels were probabilistically merged during training. For example, hepatic vessels could be merged with the liver or lesions with the surrounding tissue, encouraging the model to produce anatomically consistent segmentations across multiple levels of the label hierarchy.

\subsubsection{Training datasets}
\label{sec:trainind_dataset}

A full summary of all training datasets is provided in Table 1 of the Supplementary Material. Following prior work, we assembled the training dataset from publicly available datasets spanning multiple imaging modalities with diverse segmentation annotations. These included Computed Tomography (CT)~\cite{roth2018new, mayr2023cervical, stoverud2024aeropath, mayer2024type, MP-COVID-19-SegBenchmark, riera2025calibration, yang2017data, dap_atlas, topcowchallenge, imran2025multi, mehrad_aria_mustafa_ghaderzadeh_farkhondeh_asadi_2021, wang2024sammed3dgeneralpurposesegmentationmodels, wu2025mswal, yan2018deeplesion, deng2021ctspine1k, li2024abdomenatlas, aerts2015data, yang2021ribseg, 2024TMI, ma2024automatic}, Magnetic Resonance Imaging (MRI)~\cite{mayr2023cervical, lalande2020emidec, grovik2020deep, duke_liver, marcus2007open, wahid2024overview, pace2024hvsmr, d2024totalsegmentator, van2023spider, garrucho2025large, martin2023deep, quinton2023tumor, adams2022prostate158, LLD-MMRI}, Positron Emission Tomography (PET)~\cite{gatidis2022whole, dap_atlas}, Ultrasound (US)~\cite{xiao2016resect, leclerc2019deep, duque2024ultrasound, stanford_cine, ndzimbong2025trusted, kronke2022tracked}, and Microscopy~\cite{wei2021axonem, lin2021nucmm, zheng2023nis3d, svoboda2009generation, SELMA3D-4, ZerovnikMekuc2020, CREMI2016}.

As ultrasound is a primary application domain of our research, we further enriched the training data with two additional US datasets. The first comprised a privately collected dataset of 3D ultrasound volumes acquired from healthy volunteers at IRCAD France under institutional approval and written informed consent. The dataset includes sweeps of the liver, kidney, and hepatic vasculature acquired using a Telemed MicroUS probe equipped with an NDI Trackstar electromagnetic tracking sensor. Of these, 1,151 volumes were represented as spatio-temporal sequences, while 646 volumes were reconstructed into 3D volumes using freehand 3D reconstruction via linear interpolation. The second additional dataset, referred as FLL-US-Train comprised 555 short intra-operative ultrasound video clips collected at Humanitas Research hospital, containing focal liver lesion segmentations. These clips were represented as spatio-temporal volumes.

\subsubsection{Hyper-parameter selection}

Because of the substantial training time, no formal hyperparameter optimization procedure was conducted. Instead, the model architecture and training configuration were iteratively refined through empirical experimentation and observation during development. A small number of hyperparameters were adjusted during this process, primarily the learning rate and batch size. All other hyperparameters (including architectural configurations, optimizer settings, etc.) were specified a priori and kept fixed throughout development. All hyperparameters are provided in the supplementary material (Table 3).

\section{Method validation}
We validated SLIP using a three-part validation structure similar to nnInteractive, comprising (i) a large-scale simulated-user study, (ii) a complementary real-user study, and (iii) model ablations. The simulated-user evaluation enables controlled and standardized comparison across multiple datasets, cases, and interaction rounds, while the real-user study validates performance under realistic annotation conditions and user behavior.

\subsection{Simulated user validation}
\subsubsection{Study Design and Evaluation Metrics}

All methods were evaluated in an interactive segmentation setting following the protocol commonly adopted in interactive segmentation benchmarks \cite{liu2023simpleclick, kirillov2023segment}. User interactions were generated automatically from the disagreement between the current prediction and the reference annotation, which was treated as ground truth. The simulation began with a positive click placed randomly inside the target structure. Subsequent corrective foreground or background clicks were placed in the largest false-negative or false-positive error region, respectively. The process was repeated until a predefined interaction budget ($T_{\max}$) was reached.

The FM3BIS CVPR 2025 Challenge evaluated methods using a budget of $T_{\max}=5$ interactions. While sufficient to assess early-stage performance, such a limited budget does not fully characterize interactive annotation workflows, particularly for methods with low interaction latency, where users can provide substantially more feedback and progressively refine challenging segmentations with minimal waiting time. To better capture this regime, we increased the interaction budget to $T_{\max}=50$.

After each interaction $t \in [0, T_{\max}]$, we recorded the DSC, \emph{interaction latency}, and \emph{cumulative interaction latency}. We additionally report DC-AUC@50, defined as the area under the DSC--interaction curve over the first 50 interactions.

Interaction latency was defined as the elapsed computational time between receiving the prompts for a given interaction and obtaining the corresponding updated prediction. This included prompt-dependent data processing, model inference, and prediction post-processing, while excluding model loading and any preprocessing performed before the first interaction. Cumulative interaction latency was defined as the sum of interaction latencies up to a given interaction. This metric reflects the total waiting time experienced during an interactive segmentation session and therefore provides a practical measure of annotation efficiency.

For each dataset, we report the mean and standard deviation of each metric across all test cases. Overall performance was assessed using the \emph{cross-dataset average}, computed as the mean of the per-dataset means.

\subsubsection{Baselines, datasets, and benchmarking machine}

We compared SLIP against nnInteractive, the top performing published method in the FM3BIS CVPR 2025 Challenge, using the author-provided implementation. We additionally evaluated two competitive methods with publicly available code: SegVol and SAM-Med3D.

Due to the unavailability of the FM3BIS CVPR 2025 Challenge test dataset, evaluation was performed on 13 representative public medical image segmentation datasets spanning diverse anatomical structures and imaging modalities. CT datasets included Pengwin~\cite{liu2023pelvic}, Colorectal Liver Metastases~\cite{simpson2024preoperative}, ACC-Ki67~\cite{moawad2023voxel}, HCC Liver~\cite{moawad2023multimodality}, HCC Lesion~\cite{moawad2023multimodality}, RIDER Lung~\cite{zhao2015coffee}, TRUSTED~\cite{ndzimbong2025trusted}, LNQ~\cite{dorent2024lnq}, SegRap~\cite{luo2025segrap2023}, and TUMSeg~\cite{jensen20243d}. MRI included the HANSEG dataset~\cite{podobnik2023han}, and ultrasound included the VTUS~\cite{yang2025vivim} and UterUS~\cite{bones2024automatic} datasets. None of these datasets are included in the training data. A description of each test dataset is provided in the Supplementary Material (Table 2).

All methods were benchmarked on the same workstation (Ubuntu 24.04; AMD EPYC 7742; 512\,GB RAM; NVIDIA DGX Station with NVIDIA A100 GPUs, 40\,GB memory per GPU). To ensure fair timing measurements, methods were executed sequentially under identical conditions, with each method run independently.

\subsubsection{Results and discussion}
Overall, the simulated-user evaluation demonstrates that SLIP combines state-of-the-art segmentation performance with substantially lower interaction latency than existing methods. While nnInteractive achieves slightly higher segmentation accuracy during the earliest stages of refinement, SLIP continues to improve over a larger number of interactions, resulting in higher final segmentation accuracy together with substantially lower cumulative interaction latency.

\begin{figure}
\centering
\begin{subfigure}{0.48\textwidth}
    \centering
    \includegraphics[width=\linewidth]{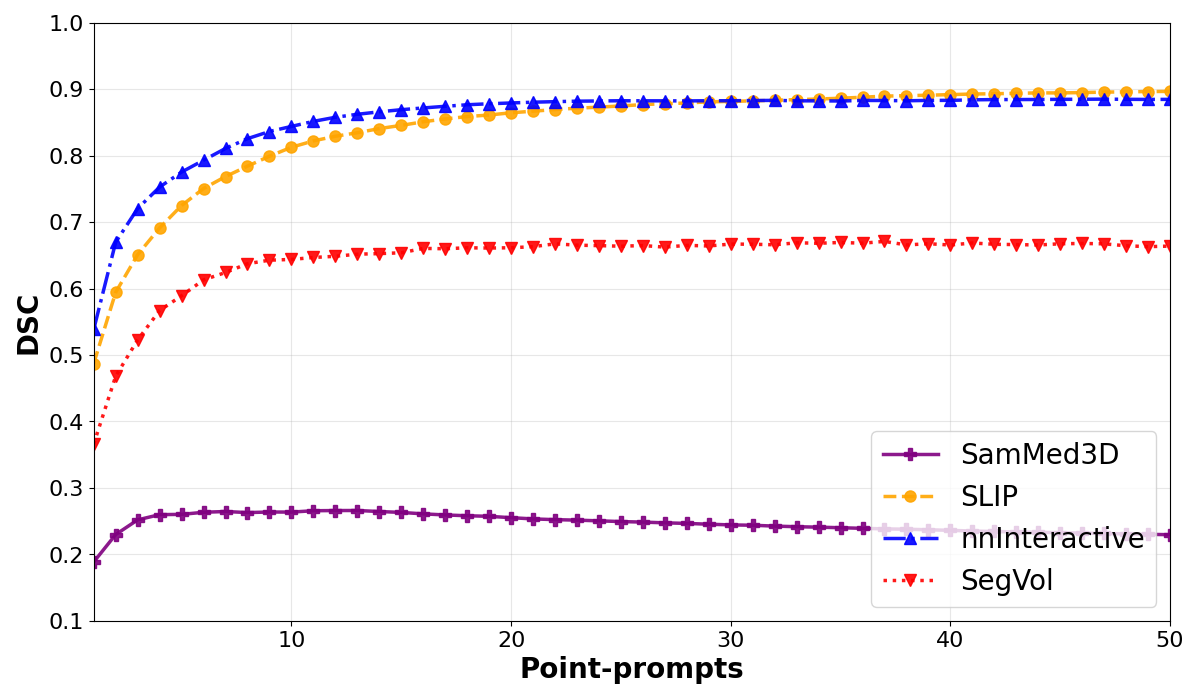}
    \caption{}
    \label{fig:dice_clicks}
\end{subfigure}
\hfill
\begin{subfigure}{0.48\textwidth}
    \centering
    \includegraphics[width=\linewidth]{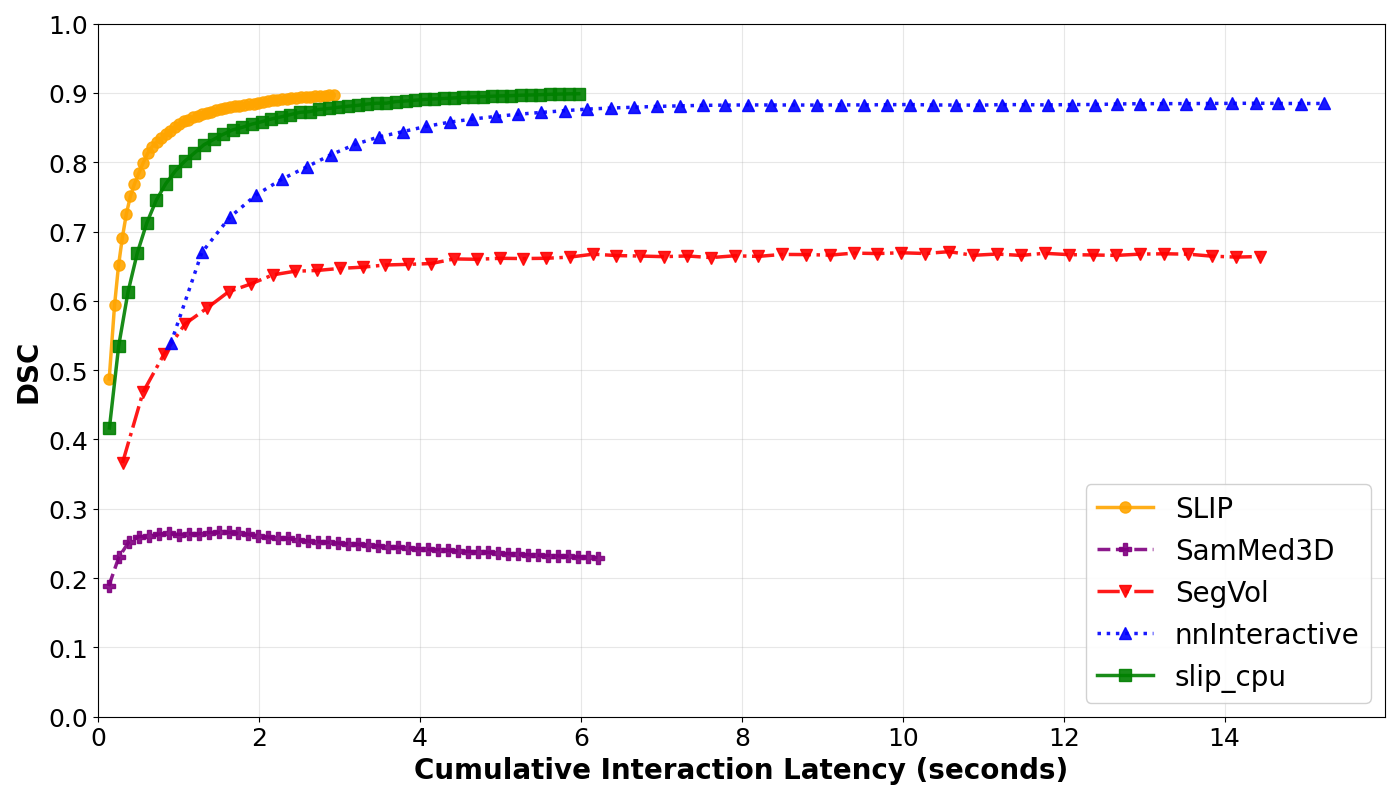}
    \caption{}
    \label{fig:dice_time}
\end{subfigure}
\caption{Quantitative comparison of segmentation performance in the simulated-user evaluation over 50 iterative interactions. (a) Mean DSC versus the number of interaction points, illustrating prompt efficiency. (b) Mean DSC versus cumulative interaction latency, illustrating the cumulative waiting time required to achieve a given segmentation quality during an interactive annotation session.}
\label{fig:interaction_comparison}
\end{figure}

Figure~\ref{fig:dice_clicks} plots the cross-dataset average DSC as a function of the number of user interactions. During the early stages of refinement ($<25$ interactions), nnInteractive achieves slightly higher DSC than SLIP. However, this advantage progressively diminishes as additional interactions are provided. A crossover occurs at approximately 30 interactions, after which SLIP consistently outperforms nnInteractive. Whereas nnInteractive exhibits a pronounced performance plateau after approximately 25 interactions, SLIP continues to improve throughout the evaluated 50-interaction budget. Importantly, segmentation quality during the earliest interactions remains substantially below the final performance achieved by either method, indicating that further refinement is often required in practice. Consequently, maintaining responsiveness to successive user interactions is likely to be more valuable than achieving a small advantage during the initial refinement steps, particularly when interaction latency is low.

Table~\ref{tab:combined_results} provides a per-dataset comparison of DSC@50, DC-AUC@50, and mean interaction latency. Across all datasets, SLIP achieves substantially lower interaction latency than all competing methods, with a cross-dataset average latency of 0.06~s per interaction compared with 0.31~s for nnInteractive, 0.29~s for SegVol, and 0.12~s for SAM-Med3D. This corresponds to approximately 5.2$\times$, 4.8$\times$, and 2.0$\times$ lower interaction latency, respectively, substantially reducing the waiting time between successive user interactions and enabling rapid iterative refinement.

In terms of segmentation accuracy, SLIP and nnInteractive consistently outperform SegVol and SAM-Med3D across all datasets. SLIP achieves the highest DSC@50 on 11 of the 13 datasets, whereas nnInteractive achieves the highest DSC@50 on 7 of the 13 datasets, with several results tied when rounded to three significant figures. Conversely, nnInteractive generally achieves higher DC-AUC@50 scores, obtaining the best result on 9 of the 13 datasets compared with 4 of the 13 for SLIP. This behaviour is consistent with Fig.~\ref{fig:dice_clicks}, indicating that nnInteractive generally converges more rapidly during the earliest interactions, whereas SLIP exhibits greater responsiveness to continued user refinement.

Figure~\ref{fig:dice_time} presents the cross-dataset average DSC as a function of cumulative interaction latency. Although SLIP initially trails nnInteractive in segmentation accuracy, its substantially lower interaction latency enables users to achieve comparable—and eventually superior—segmentation quality while waiting only a fraction of the cumulative interaction latency. After 50 interactions, the cumulative interaction latency of nnInteractive is 15.2~s compared with only 2.9~s for SLIP, highlighting the practical advantage of low-latency interaction during prolonged annotation sessions.

\begin{figure*}[t]
\centering
\setlength{\tabcolsep}{2pt}
\renewcommand{\arraystretch}{1}

\begin{tabular}{c c c c c c}
\textbf{} & \textbf{GT} & \textbf{SegVol} & \textbf{SAMMed3D} & \textbf{nnInteractive} & \textbf{SLIP} \\

\multirow{2}{*}{\textbf{HCC Liver}} 
& \includegraphics[width=0.15\textwidth]{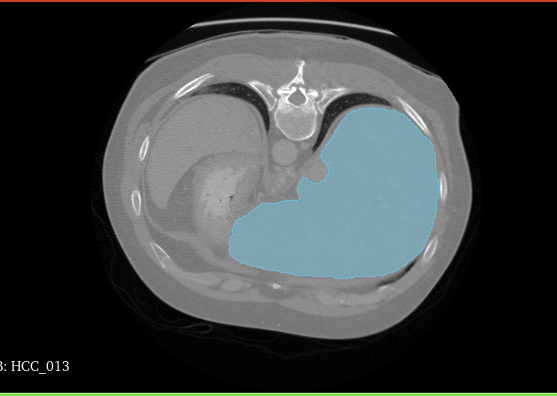} 
& \includegraphics[width=0.15\textwidth]{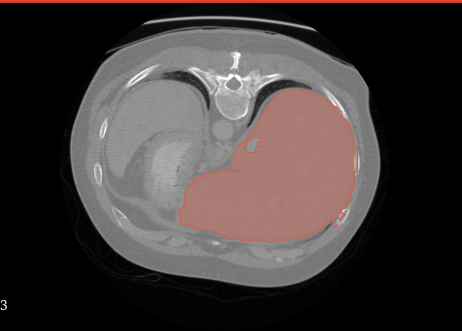} 
& \includegraphics[width=0.15\textwidth]{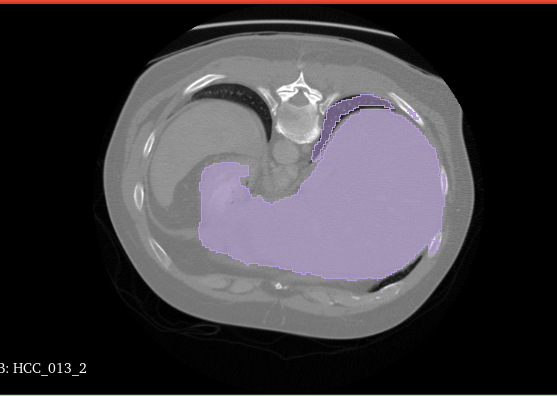} 
& \includegraphics[width=0.15\textwidth]{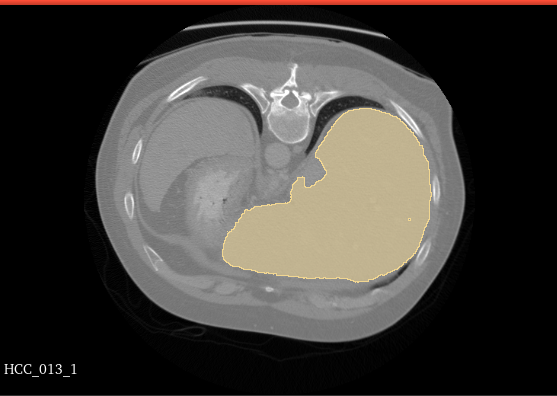} 
& \includegraphics[width=0.15\textwidth]{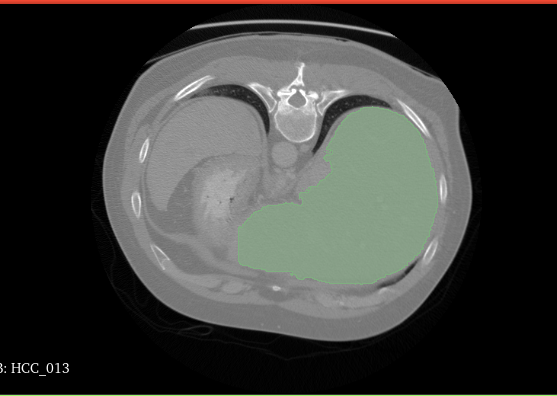} \\

& \includegraphics[width=0.15\textwidth]{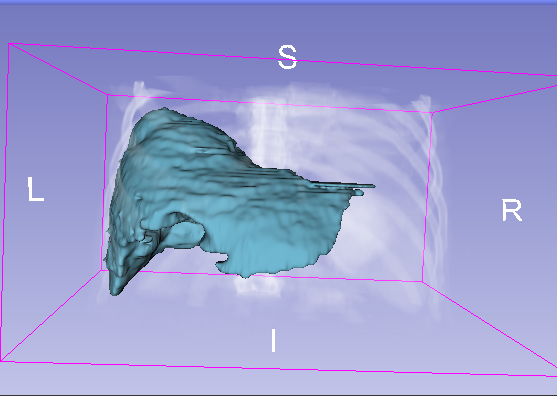} 
& \includegraphics[width=0.15\textwidth]{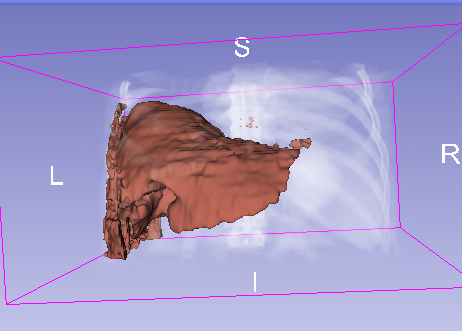} 
& \includegraphics[width=0.15\textwidth]{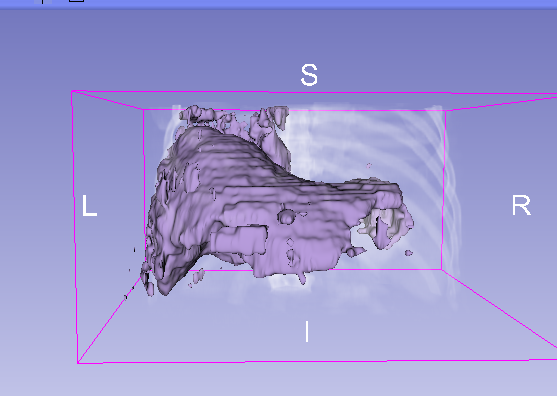} 
& \includegraphics[width=0.15\textwidth]{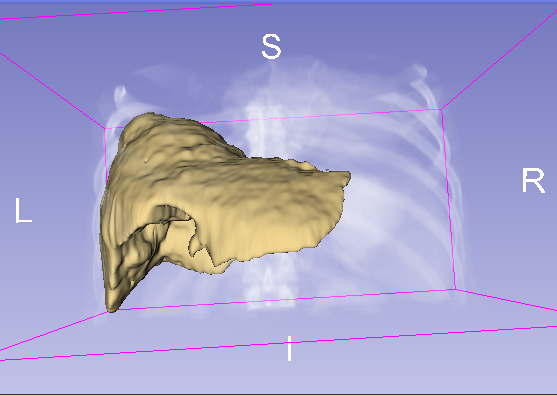} 
& \includegraphics[width=0.15\textwidth]{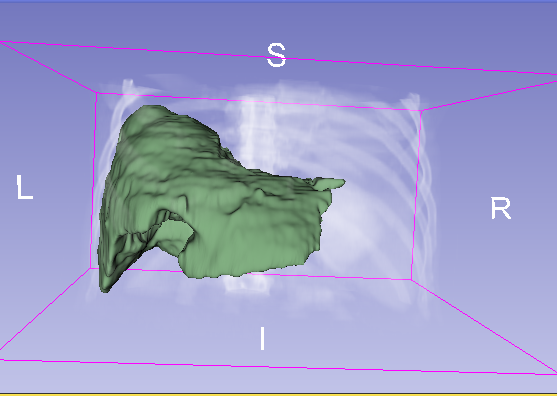} \\
\

\multirow{2}{*}{\textbf{UterUS}} 
& \includegraphics[width=0.15\textwidth]{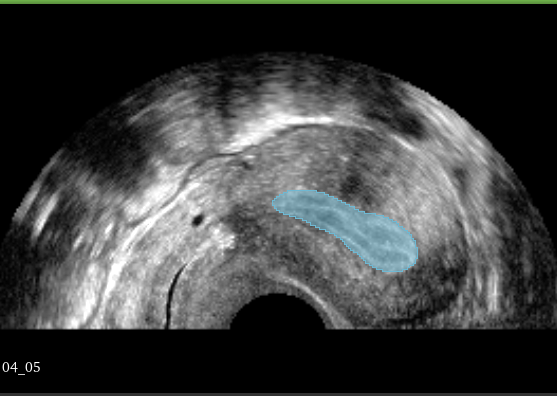} 
& \includegraphics[width=0.15\textwidth]{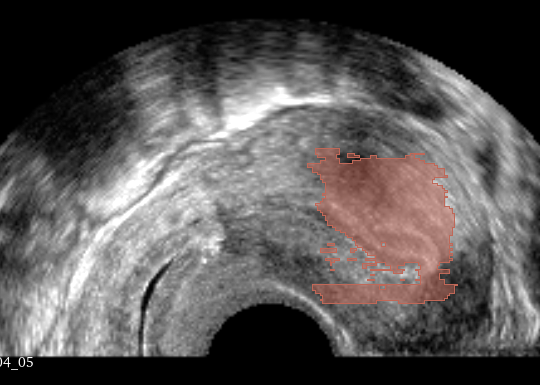} 
& \includegraphics[width=0.15\textwidth]{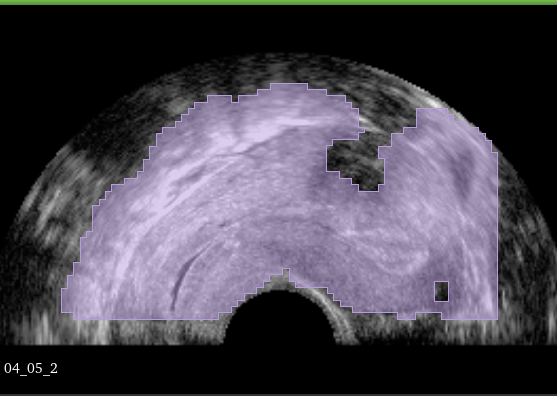} 
& \includegraphics[width=0.15\textwidth]{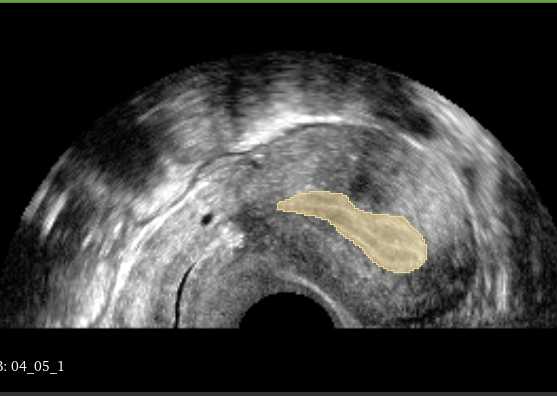} 
& \includegraphics[width=0.15\textwidth]{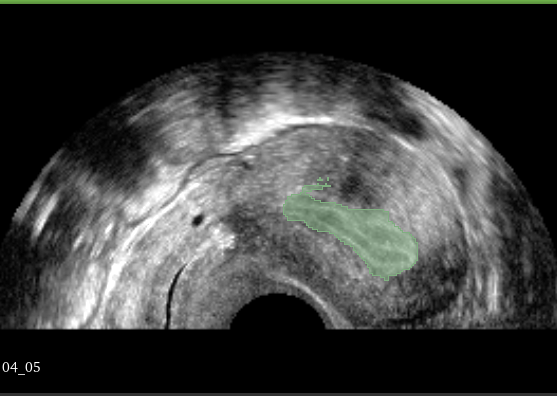} \\

& \includegraphics[width=0.15\textwidth]{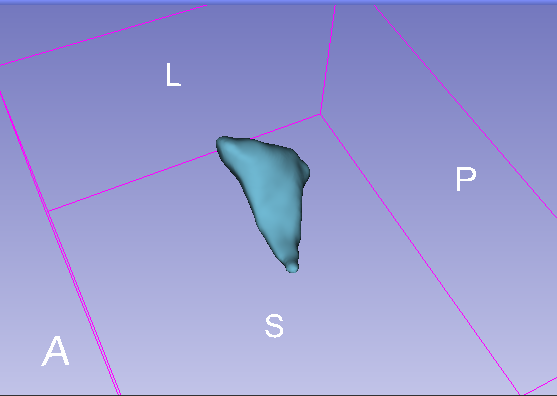} 
& \includegraphics[width=0.15\textwidth]{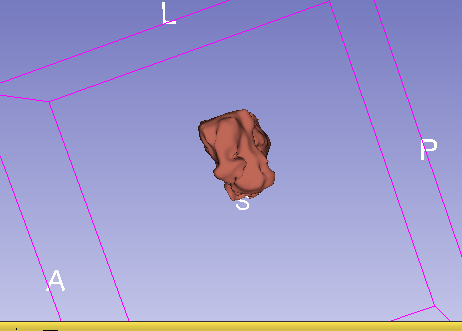} 
& \includegraphics[width=0.15\textwidth]{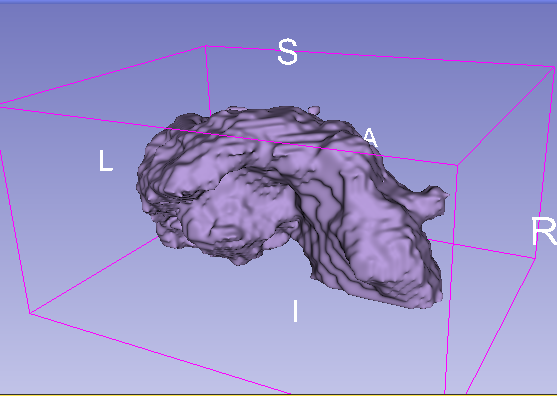} 
& \includegraphics[width=0.15\textwidth]{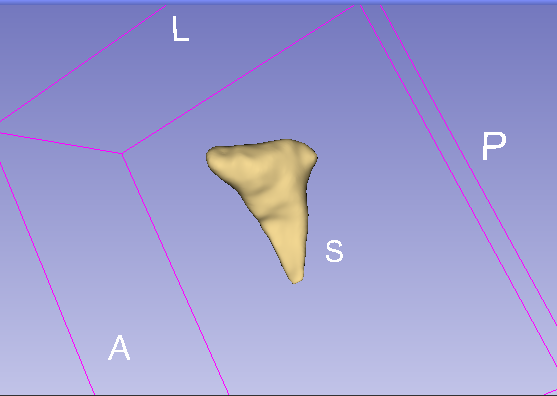} 
& \includegraphics[width=0.15\textwidth]{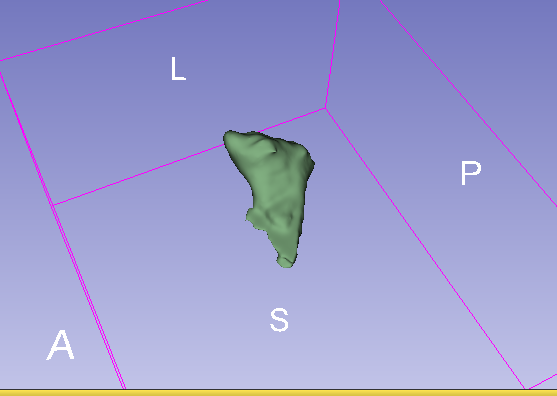} \\

\multirow{2}{*}{\textbf{Pengwin}} 
& \includegraphics[width=0.15\textwidth]{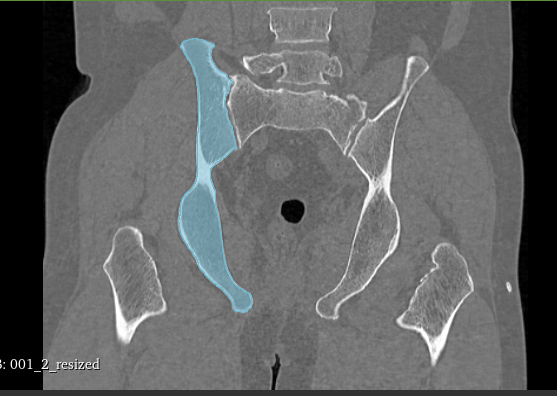} 
& \includegraphics[width=0.15\textwidth]{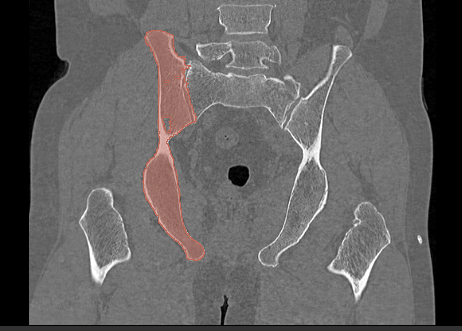} 
& \includegraphics[width=0.15\textwidth]{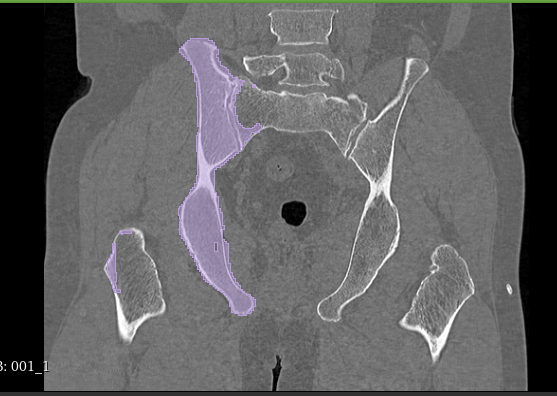} 
& \includegraphics[width=0.15\textwidth]{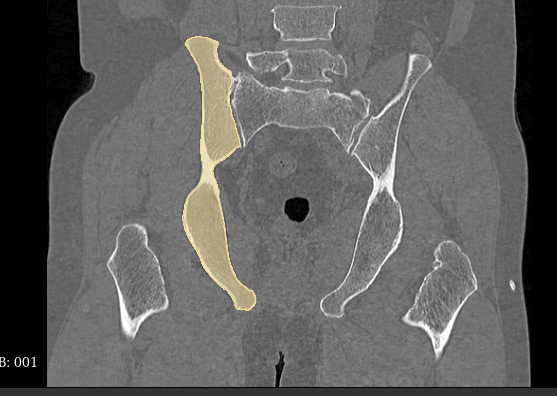} 
& \includegraphics[width=0.15\textwidth]{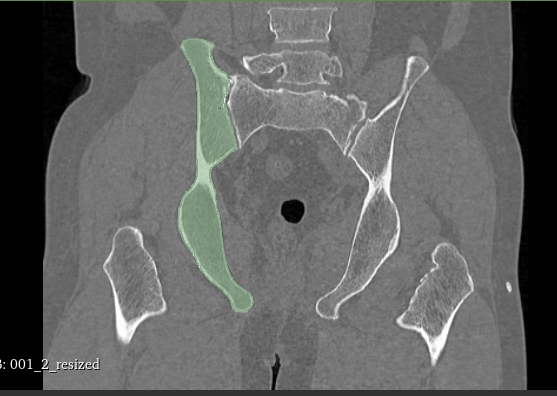} \\

& \includegraphics[width=0.15\textwidth]{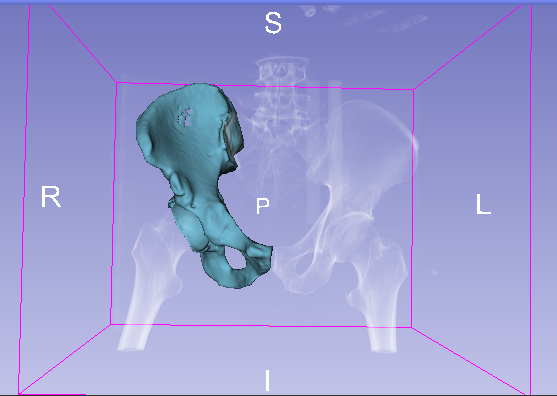} 
& \includegraphics[width=0.15\textwidth]{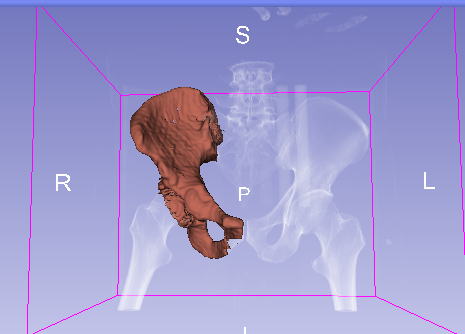} 
& \includegraphics[width=0.15\textwidth]{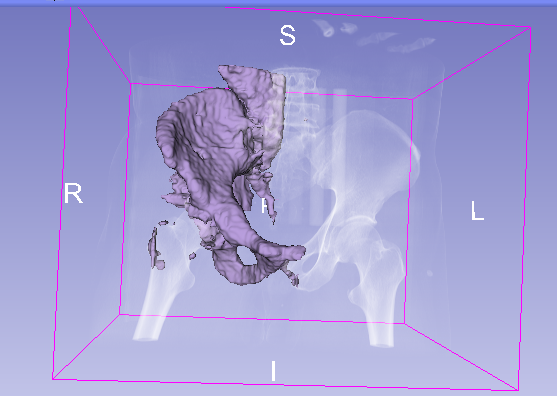} 
& \includegraphics[width=0.15\textwidth]{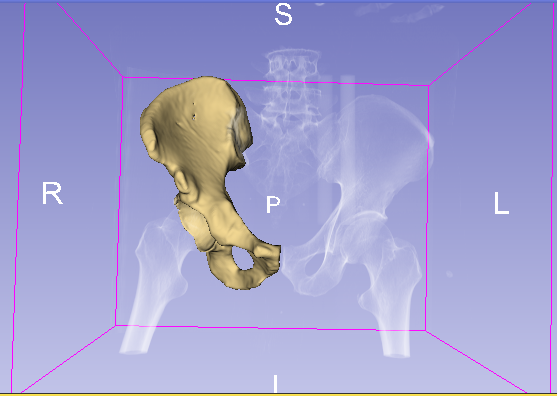} 
& \includegraphics[width=0.15\textwidth]{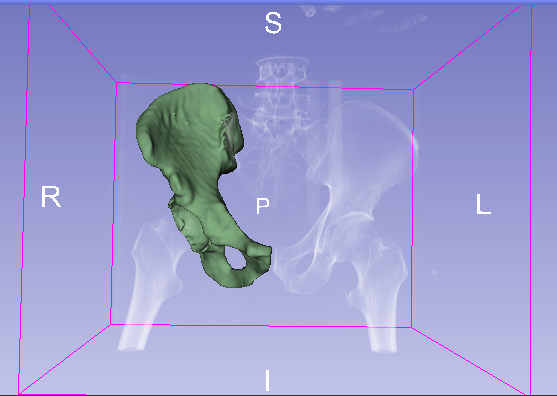} \\

\end{tabular}

\caption{Qualitative comparison across three representative test datasets in the simulated-user evaluation. Each row shows a representative case, with a 2D image slice (top) and corresponding 3D rendering (bottom). Columns show the ground truth (GT) and segmentations produced by SegVol, SAMMed3D, nnInteractive, and SLIP after 10 simulated iterative clicks.}
\label{fig:qualitative_results}
\end{figure*}

Finally, Fig.~\ref{fig:qualitative_results} presents representative qualitative examples after ten interactions. SegVol and SAM-Med3D frequently fail to recover accurate object boundaries or preserve spatial consistency across anatomically challenging structures, whereas both nnInteractive and SLIP produce substantially more accurate and coherent segmentations. In particular, the examples highlight the ability of SLIP and nnInteractive to preserve fine structural detail and recover complex anatomical boundaries that are frequently missed by SegVol and SAM-Med3D.

Overall, these findings indicate that interaction latency and sustained responsiveness to successive user corrections are important complementary evaluation criteria to segmentation accuracy when assessing practical interactive segmentation systems. While nnInteractive provides stronger performance during the earliest refinement stages, SLIP achieves superior final segmentation accuracy together with substantially lower cumulative interaction latency, supporting efficient prolonged interactive refinement.

\paragraph{CPU deployment}
We additionally evaluated SLIP in a CPU-only deployment setting to assess its suitability for environments where GPU resources are unavailable during interactive annotation. Because image encoding is decoupled from prompt-guided refinement, image features can be precomputed once before interaction on either a CPU or GPU. In our experiments, image features were precomputed on a GPU to isolate interactive performance, after which all prompt-guided inference was performed on the CPU. To maintain low interaction latency, patch feature contextualization via the memory bank was disabled. As shown in Fig.~\ref{fig:dice_time}, SLIP achieved a mean interaction latency of 0.12~s per interaction during CPU inference while preserving competitive segmentation accuracy. This remained substantially faster than nnInteractive, which required approximately 15~s per interaction under the same deployment setting. 

To the best of our knowledge, no existing state-of-the-art 3D interactive segmentation method currently provides comparably low interaction latency during CPU-based interactive segmentation. These results suggest that decoupling image encoding from prompt-guided refinement enables practical interactive deployment in resource-constrained environments where GPU acceleration during annotation is unavailable.

\begin{figure}
    \centering
    \includegraphics[width=1.0\linewidth]{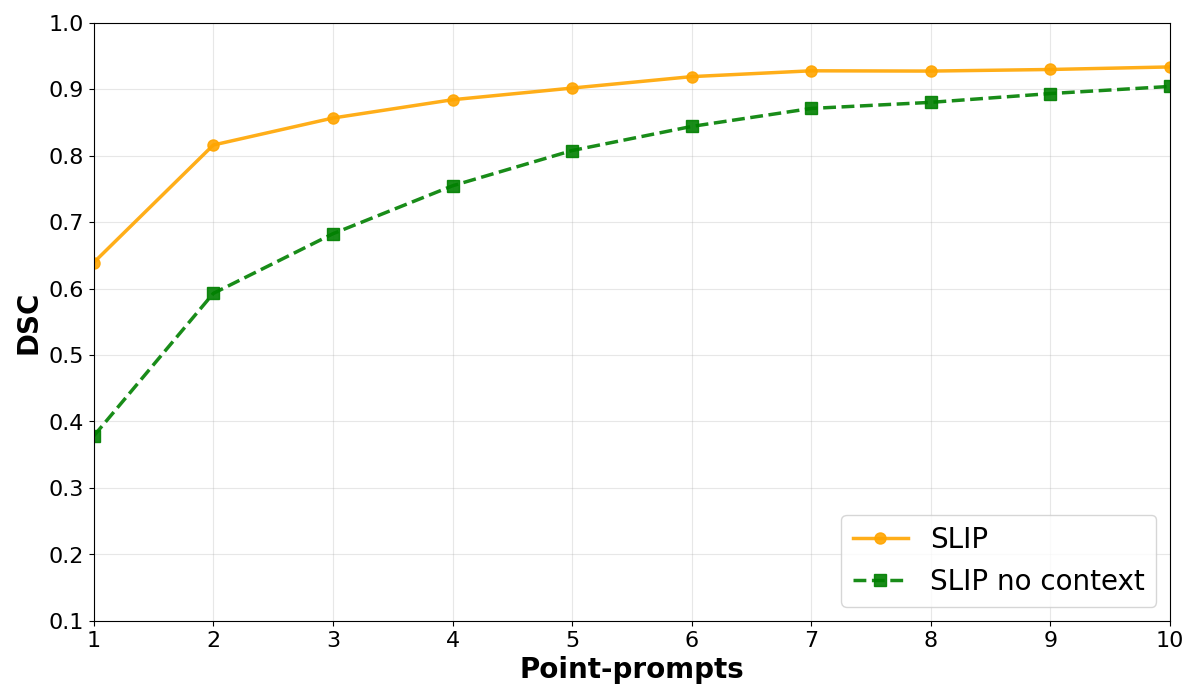}
    \caption{Ablation of patch feature contextualization for simulated interactive liver segmentation. Mean DSC is plotted against the number of interaction points for models with and without patch feature contextualization, illustrating its role in propagating sparse user guidance during the earliest stages of refinement.}
    \label{fig:propagation_liver}
\end{figure}

\begin{table*}[t]
\centering
\small
\begin{tabular}{l|ccc|ccc|ccc|ccc}
\hline
 & \multicolumn{3}{c|}{SLIP} 
 & \multicolumn{3}{c|}{nnInteractive} 
 & \multicolumn{3}{c|}{SegVol} 
 & \multicolumn{3}{c}{SAM-Med3D} \\
Dataset 
 & Final & AUC & Time 
 & Final & AUC & Time 
 & Final & AUC & Time 
 & Final & AUC & Time \\
\hline
HANSEG        & \textbf{0.78} & \textbf{0.70} & \textbf{0.04} & 0.74 & 0.69 & 0.20 & 0.45 & 0.43 & 0.15 & 0.09 & 0.09 & 0.12 \\
PENGWIN       & 0.94 & 0.90 & \textbf{0.10} & \textbf{0.95} & \textbf{0.95} & 0.39 & 0.75 & 0.74 & 0.67 & 0.26 & 0.26 & 0.13 \\
Colorectal-LM & \textbf{0.86} & \textbf{0.80} & \textbf{0.05} & 0.81 & 0.76 & 0.28 & 0.53 & 0.53 & 0.14 & 0.05 & 0.05 & 0.12 \\
ACC-KI67      & \textbf{0.93} & 0.91 & \textbf{0.07} & \textbf{0.93} & \textbf{0.93} & 0.28 & 0.76 & 0.79 & 0.23 & 0.38 & 0.38 & 0.13 \\
HCC-Liver     & \textbf{0.96} & 0.94 & \textbf{0.05} & \textbf{0.96} & \textbf{0.95} & 0.40 & 0.91 & 0.89 & 0.65 & 0.72 & 0.72 & 0.14 \\
HCC-Lesion    & \textbf{0.86} & 0.79 & \textbf{0.06} & 0.85 & \textbf{0.81} & 0.35 & 0.61 & 0.59 & 0.16 & 0.21 & 0.21 & 0.13 \\
RIDER-Lung    & \textbf{0.90} & 0.86 & \textbf{0.07} & 0.89 & \textbf{0.87} & 0.34 & 0.72 & 0.70 & 0.16 & 0.06 & 0.06 & 0.13 \\
TRUSTED       & \textbf{0.97} & 0.96 & \textbf{0.08} & \textbf{0.97} & \textbf{0.97} & 0.43 & 0.88 & 0.86 & 0.50 & 0.71 & 0.71 & 0.13 \\
TUMSeg        & \textbf{0.94} & 0.90 & \textbf{0.04} & 0.94 & \textbf{0.92} & 0.26 & 0.74 & 0.70 & 0.21 & 0.02 & 0.02 & 0.11 \\
UterUS        & 0.91 & 0.79 & \textbf{0.03} & \textbf{0.92} & \textbf{0.88} & 0.23 & 0.58 & 0.56 & 0.16 & 0.08 & 0.08 & 0.09 \\
VTUS          & \textbf{0.90} & \textbf{0.87} & \textbf{0.02} & 0.88 & 0.85 & 0.42 & 0.61 & 0.57 & 0.42 & 0.40 & 0.40 & 0.10 \\
LNQ           & \textbf{0.87} & 0.82 & \textbf{0.06} & \textbf{0.87} & \textbf{0.86} & 0.25 & 0.65 & 0.64 & 0.14 & 0.12 & 0.12 & 0.13 \\
SEGRAP        & \textbf{0.83} & \textbf{0.75} & \textbf{0.04} & 0.78 & 0.73 & 0.34 & 0.41 & 0.40 & 0.15 & 0.12 & 0.12 & 0.13 \\
\hline
\textbf{Average} 
              & \textbf{0.90} & 0.85 & \textbf{0.06}
              & 0.88 & \textbf{0.86} & 0.31
              & 0.66 & 0.65 & 0.29
              & 0.25 & 0.25 & 0.12 \\
\end{tabular}

\caption{Dataset-wise comparison including final DSC (after 50 clicks), DSC AUC, and mean interaction time per click (seconds). Best performance per metric is highlighted in bold.}
\label{tab:combined_results}
\end{table*}
\subsection{Prospective User Study}

While simulated-user evaluation provides a scalable and standardized benchmark for segmentation performance, it cannot fully assess practical interactive annotation workflows. Automatically generated interactions cannot capture human decision-making, error-correction strategies, adaptation to intermediate predictions, or subjective aspects of usability such as perceived responsiveness and the benefits of reversible prompting. We therefore complemented the simulated evaluation with a prospective controlled user study to assess annotation efficiency and user experience under realistic interactive annotation conditions.

\subsubsection{Annotation Tasks}

Three annotation tasks were evaluated.

The \textbf{MRI Uterus Task} comprised 15 randomly selected MRI volumes from the external public UT-EndoMRI dataset~\cite{liang2025multi}. Participants segmented the uterus in each volume (one volume per patient).

The \textbf{CT Liver Lesion Task} comprised 15 randomly selected CT volumes from the external public IRCADB1 dataset~\cite{soler20103d}. Participants segmented liver lesions in each volume (one volume per patient).

The \textbf{Ultrasound Liver Lesion Task} comprised 15 short intraoperative ultrasound video clips (mean duration 4.0~s, standard deviation 1.9~s) drawn from the held-out patient test split of the FLL-US-Train dataset. Unlike the MRI and CT tasks, this evaluation was intentionally designed as a task-supervised evaluation, allowing the model to learn from task-specific training data before being evaluated on held-out patients. This reflects the practical scenario in which a general interactive segmentation framework is subsequently retrained or fine-tuned on a newly available task-specific dataset before being used to annotate additional cases. Participants segmented all focal liver lesions visible in each clip (1--3 lesions per clip; one clip per patient). To isolate segmentation performance from lesion detection, an arrow was placed at a random location within each target lesion to indicate the lesion of interest. The same arrow locations were used for all participants.

Although the public datasets include reference annotations, the annotation protocols used to generate them are not publicly available, leaving important aspects of the annotation specification undefined. A study-specific annotation protocol was therefore developed to ensure consistent annotations across participants.

\subsubsection{Participants, compared annotation methods, and study end-points}

The study included six participants: three radiographers with professional medical image annotation experience and three junior surgeons. Owing to resource constraints, senior radiologists and experienced surgeons were not included; therefore, the findings may not fully generalize to these user groups. Nevertheless, the cohort is representative of many medical image AI annotation workflows, where trained annotators and junior clinicians contribute substantially to dataset annotation.

Each participant independently completed all segmentation tasks using three annotation methods in a within-subject repeated-measures design, enabling direct comparison while controlling for inter-user variability. All methods were implemented within 3D Slicer:

\begin{enumerate}
\item Manual segmentation using the built-in Segmentation Editor (manual), a widely used medical image annotation tool.
\item Interactive segmentation with nnInteractive, selected as the strongest baseline. The 3D Slicer plugin described in~\cite{de2025slicernninteractive} was used.
\item Interactive segmentation with SLIP, implemented within the same plugin framework and extended to support reversible prompting through native prompt undo.
\end{enumerate}

Participants were instructed to produce the most accurate segmentation possible, and no explicit time limit was imposed. An appropriate limit could not be determined \emph{a priori}, and enforcing one risked leaving challenging cases only partially refined, thereby confounding comparisons between methods. Because nnInteractive does not support prompt undo, participants wishing to revise an interaction were permitted to restart the annotation from scratch. Restarts were limited to three per case to allow recovery from occasional user errors while preventing unlimited retries when satisfactory segmentations could not be achieved.

The primary endpoints were \emph{annotation time}, \emph{annotation reference consistency}, and \emph{user-reported interaction quality}. User-reported interaction quality was assessed using post-task questionnaires evaluating perceived latency, responsiveness, ability to correct segmentation errors, and overall model preference. Secondary endpoints included the number of interaction points, the number of undo actions when using SLIP, and the number of annotation restarts when using nnInteractive.
\subsubsection{Study design and statistical analysis}

As with any prospective user study, several factors could influence the validity of the findings. These were considered during study design and addressed through a combination of experimental controls, standardized study procedures, and statistical analysis (Table~\ref{tab:userstudy_biases}).

\begin{table}[t]
\centering
\small
\begin{tabularx}{\linewidth}{>{\raggedright\arraybackslash\bfseries}p{0.30\linewidth}X}
\toprule
Potential bias & Mitigation strategy \\
\midrule

Participant selection &
Participants included both professional medical image annotators and junior surgeons. Although senior radiologists and experienced surgeons were not included owing to resource constraints, the cohort is representative of many medical image AI annotation workflows. \\

Image selection &
One image per patient was randomly selected from each source dataset. \\

Case difficulty &
A repeated-measures design was used in which every participant annotated every image using every annotation method. \\

Software interface &
All annotation methods were implemented within 3D Slicer using a consistent user interface and comparable interactive plugins. \\

Annotation protocol and software familiarity &
Participants followed a study-specific annotation protocol and completed an individual hour orientation session, lasting approximately 45 - 60 mins, including three practice cases per task using each annotation method. The study commenced only after participants confirmed familiarity with both the annotation protocol and software. \\

Ordering, learning, and context switching &
The order of annotation methods was randomized independently for each participant. Images were presented in method-specific blocks to reduce context switching, and schedules were balanced so that each method was equally likely to be performed first, second, or third. \\

Image interpretation time &
Each case comprised an untimed image inspection phase followed by a timed annotation phase. \\

Participant fatigue &
Participants were instructed to take a 5-minute break after every 20 minutes of annotation. \\

Questionnaire interpretation &
Questionnaire items used neutral Likert-scale wording and were independently reviewed by an experienced medical image annotation professional to improve clarity and minimize ambiguity. \\

\bottomrule
\end{tabularx}
\caption{Potential sources of bias in the prospective user study and corresponding mitigation strategies.}
\label{tab:userstudy_biases}
\end{table}

The study design reduces, but does not eliminate, variability arising from participant ability, intrinsic case difficulty, and repeated measurements. To account for the remaining variability, annotation times were log-transformed prior to analysis to reduce right-skewness and better satisfy model assumptions. Because the three segmentation tasks differed substantially in imaging modality, anatomy, and baseline annotation time, separate linear mixed-effects models (LMMs) were fitted for each task:

\begin{equation}
\log(\mathrm{Time}) \sim \mathrm{Method} + \mathrm{Difficulty} + (1 \mid \mathrm{User}) + (1 \mid \mathrm{Volume}).
\end{equation}

Here, \emph{Method} (manual segmentation as the reference level) and participant-reported \emph{Difficulty}  were included as fixed effects. Crossed random intercepts for \emph{User} and \emph{Volume} accounted for the repeated-measures design and residual differences between participants and images. Effects are reported as percentage changes in annotation time relative to manual segmentation, together with 95\% confidence intervals and corresponding $p$-values.

\paragraph{Measuring annotation reference consistency}

Study-specific reference annotations were generated because the public datasets do not document the annotation protocols or quality-control procedures used to produce their labels. Consequently, agreement with the public reference annotations could reflect differences in annotation protocol rather than differences between annotation methods.

Reference annotations were generated from the six manual segmentations for each image using the STAPLE algorithm~\cite{warfield2004simultaneous}, which estimates a probabilistic consensus segmentation by modelling each annotator's sensitivity and specificity. Only the manual segmentations were used to construct the consensus to avoid biasing the reference toward either interactive method.

Annotation reference consistency was quantified by computing the Dice similarity coefficient (DSC) between each SLIP or nnInteractive segmentation and the corresponding STAPLE consensus segmentation. Task-level comparisons were performed using paired differences in DSC between SLIP and nnInteractive. Because these differences were not normally distributed, statistical significance was assessed using a two-sided Wilcoxon signed-rank test with a significance threshold of $p<0.05$.

\paragraph{Study hardware}

Participants were recruited from two sites: IRCAD France (Strasbourg; $n=3$) and IRCAD Africa (Kigali; $n=3$). Both sites used comparable workstations. Participants in Strasbourg completed the study using a Linux workstation equipped with an Intel Core i9-10980 processor, 32~GB RAM, and an NVIDIA GeForce RTX~3080 GPU. Participants in Kigali completed the study using a workstation equipped with an Intel Core i9-9900K processor, 32~GB RAM, and an NVIDIA GeForce RTX~4070 GPU. Manual annotation was performed using a stylus and Wacom pen display (Wacom Cintiq in Strasbourg and Wacom MobileStudio Pro~16 in Kigali).

\paragraph{Post-task questionnaires and statistical analysis}

Questionnaires were administered at three levels throughout the study: image-level, task-level, and study-level.

\textbf{Image-level:} After annotating each image, participants rated perceived case difficulty on a five-point Likert scale (1 = easy, 5 = very difficult). These ratings were included as the \emph{VolumeDifficulty} fixed effect in the linear mixed-effects models.

\textbf{Task-level:} After completing each segmentation task, participants completed a 10-item questionnaire using seven-point Likert scales. The questionnaire evaluated four aspects of interactive segmentation: perceived latency, responsiveness, ability to correct segmentation errors, and overall model preference. The complete questionnaire is provided in Supplementary Material (Table 5).

To compare annotation methods, responses were first averaged within each thematic category (Latency, Responsiveness, Ability to Correct, and Model Preference) for each participant. Category means were then macro-averaged across participants. Owing to the small sample size and the descriptive nature of the questionnaire, responses were summarized descriptively using mean Likert scores rather than formal hypothesis testing.

\textbf{Study-level:} After completing all segmentation tasks, participants completed a short end-of-study questionnaire assessing the relative importance of key interaction characteristics for practical annotation workflows, including model latency, support for iterative refinement, and support for reversible prompting. For clarity, reversible prompting was described as the ability to remove previously placed prompts (i.e., prompt undo) without restarting the annotation. The complete questionnaire is provided in Supplementary Material (Table 6).

\subsubsection{Results and discussion}
\paragraph{Annotation time}
Across all three segmentation tasks, both interactive methods substantially reduced annotation time relative to manual segmentation, demonstrating the efficiency advantage of interactive segmentation over fully manual annotation (Table~\ref{tab:merged_user_study}), demonstrating the efficiency advantage of interactive segmentation over fully manual annotation. For the Ultrasound Liver Lesion and CT Liver Lesion tasks, SLIP and nnInteractive achieved comparable reductions in annotation time, whereas nnInteractive provided a larger reduction for the MRI Uterus task.

For the US Liver Lesion Task, the LMM estimated a 79.0\% reduction in annotation time for SLIP relative to manual segmentation ($p < 0.001$, 95\% CI: 75.2\%--82.2\%) and an 80.1\% reduction for nnInteractive ($p < 0.001$, 95\% CI: 76.5\%--83.2\%). There was no significant difference between the two interactive methods, with SLIP requiring 5.6\% longer annotation time on average ($p = 0.520$, 95\% CI: $-10.6\%$ to $+24.9\%$). Despite requiring more interaction points than nnInteractive, SLIP achieved comparable annotation times, suggesting that its lower interaction latency largely compensated for the additional refinement interactions.

For the MRI Uterus Task, the model estimated a 23.9\% reduction in annotation time for SLIP relative to manual segmentation ($p = 0.0001$, 95\% CI: 13.6\%--33.0\%) and a 66.6\% reduction for nnInteractive ($p = 0.0001$, 95\% CI: 62.0\%--70.6\%). Relative to SLIP, nnInteractive reduced annotation time by 56.0\% ($p < 0.001$, 95\% CI: 50.1.0\%--61.3\%). This difference was accompanied by substantially fewer refinement interactions, indicating that the current SLIP model required considerably more user correction for this task. Consequently, the lower interaction latency of SLIP was insufficient to offset the increased interaction burden. 

For the CT Liver Lesion Task, the model estimated a 62.3\% reduction in annotation time for SLIP ($p < 0.001$, 95\% CI: 55.6\%--68.0\%) and a 62.8\% reduction for nnInteractive ($p < 0.001$, 95\% CI: 56.1\%--68.4\%). The estimated difference between the two interactive methods was not statistically significant ($p = 0.89$, 95\% CI: [-16.1\%, 16.6\%]). Although the average annotation times were comparable, nnInteractive exhibited substantially greater variability (Table~\ref{tab:merged_user_study}). This variability was primarily attributable to 3 volumes for which repeated user interactions failed to recover an acceptable segmentation despite up to three annotation restarts, resulting in annotation times of up to 2961 seconds. In these cases, the predicted segmentation consistently expanded into the surrounding liver parenchyma, preventing accurate refinement through additional prompts. No comparable failures were observed with SLIP, for which the maximum annotation time was 1979 seconds. These observations suggest that reversible prompting enabled a more robust interaction workflow by allowing users to recover from intermediate interaction errors without discarding previous annotation progress. 

\paragraph{Annotation reference consistency and number of points}

Annotation reference consistency broadly reflected the annotation time results (Table~\ref{tab:merged_user_study}) while providing additional insight into the interaction strategies required by each method.

For the US Liver Lesion Task, both interactive methods achieved comparable agreement with the reference annotations. Although nnInteractive obtained a slightly higher mean DSC than SLIP (0.834 versus 0.823), the difference was not statistically significant ($p = 0.489$). SLIP required more interaction points than nnInteractive; however, this did not translate into longer annotation times, suggesting that its lower interaction latency largely compensated for the additional refinement interactions.

For the MRI Uterus Task, nnInteractive achieved significantly higher agreement with the reference annotations than SLIP ($p < 0.001$). This was accompanied by approximately four times fewer interaction points than required by SLIP, indicating that substantially less refinement was needed to obtain satisfactory segmentations. Nevertheless, annotation time increased by a considerably smaller factor, demonstrating that the lower interaction latency of SLIP partially offset the increased interaction burden.

For the CT Liver Lesion Task, SLIP achieved a higher mean DSC than nnInteractive (0.827 versus 0.763), although the difference was not statistically significant ($p = 0.804$). A similar number of interaction points was required for both methods, indicating that differences in annotation behaviour were not explained by interaction frequency alone. Inspection of the individual annotations revealed marked differences in refinement behaviour. Outlier analysis based on the DSC interquartile range identified three outlier cases for nnInteractive and none for SLIP. These corresponded to the same cases associated with prolonged annotation times, indicating that failures of iterative refinement, rather than an increased number of user interactions, were responsible for the observed variability in annotation performance.

\paragraph{Reversible prompting}

Annotation restarts were required in all three tasks when using nnInteractive, reflecting the absence of a mechanism to remove incorrect prompts once they had been placed. The highest number of restarts per image occurred in the CT Liver Lesion Task 0.61, where users frequently restarted the annotation after unsuccessful refinement attempts. The restart number was substantially lower for the MRI Uterus Task (0.04), indicating that the impact of irreversible prompting was task-dependent.

In contrast, SLIP supports reversible prompting through a native prompt-undo mechanism, which was used across all three tasks. The mean number of undo operations per image was 0.68, 0.76, and 1.23 for the US Liver Lesion, MRI Uterus, and CT Liver Lesion tasks, respectively. Rather than restarting an annotation, participants were able to remove incorrect prompts while preserving previous interactions and continue refining the existing segmentation. Although reversible prompting was introduced primarily as an error recovery mechanism, its consistent use across all tasks indicates that participants naturally incorporated it into their annotation workflow. Together with the reduced variability in annotation time observed for SLIP, these findings suggest that reversible prompting improves workflow robustness by enabling users to recover from intermediate interaction errors without discarding previous annotation progress.

\paragraph{Questionnaire analysis}

For the US Liver Lesion and CT Liver Lesion tasks, SLIP consistently received higher ratings across all four questionnaire categories. In contrast, for the MRI Uterus Task, nnInteractive received similar or higher ratings, consistent with its superior segmentation performance for this task.

The end-of-study questionnaire provided further insight into the interaction characteristics that participants considered most important in practical annotation workflows. Participants assigned high importance to low model latency (mean 6.3/7), preferred workflows that supported iterative refinement even when the initial prediction was imperfect (mean 6.6/7), and unanimously rated support for reversible prompting (i.e., the ability to remove previously placed prompts without restarting the annotation) as essential (mean 7.0/7).

These responses are consistent with the quantitative findings. Although SLIP generally required more interaction points than nnInteractive, its lower interaction latency enabled comparable annotation times for the liver lesion tasks, while reversible prompting eliminated the need to restart annotations following incorrect interactions. Together, these findings suggest that practical usability depends not only on segmentation accuracy, but also on interaction design, particularly low-latency interaction and the ability to efficiently recover from intermediate prediction errors.

\begin{table*}[t]
\centering
\small
\begin{tabular}{l l|ccc}
\hline
Model & Metric & US Liver Lesion Task & MRI Uterus Task & CT Liver Lesion Task \\
\hline

\multirow{1}{*}{Manual}
& Time (s)
& 451.13 $\pm$ 310.32
& 165.35 $\pm$ 106.68
& 431.46 $\pm$ 510.65 \\
\hline

\multirow{8}{*}{nnInteractive}
& Time (s)
& 100.96 $\pm$ 85.47
& \textbf{52.36 $\pm$ 28.86}
& 265.75 $\pm$ 523.88 \\
& DSC
& 0.834 $\pm$ 0.06
& \textbf{0.770 $\pm$ 0.10}
& 0.763 $\pm$ 0.19 \\
& Number of Points per image
& \textbf{16.59}
& \textbf{8.02}
& 34.7 \\
&Number of restarts per image
& 0.09
& 0.04
& 0.61 \\

\cline{2-5}
& \multicolumn{4}{l}{\textit{Questionnaire}} \\
\cline{2-5}
& Latency
& 6.39
& 6.61
& 5.39 \\
& Responsiveness
& 5.67
& \textbf{6.00}
& 5.67 \\
& Ability to Correct
& 5.00
& \textbf{5.78}
& 4.67 \\
& Model Preference
& 6.00
& \textbf{6.33}
& 5.50 \\
\hline

\multirow{9}{*}{SLIP}
& Time (s)
& \textbf{100.28 $\pm$ 78.49}
& 131.23 $\pm$ 74.92
& \textbf{224.48 $\pm$ 381.06} \\
& DSC
& 0.832 $\pm$ 0.05
& 0.730 $\pm$ 0.09
& \textbf{0.827 $\pm$ 0.06} \\
& Number of Points per image
& 21.93
& 36.77
& \textbf{33.6} \\
& Number of prompt undos per image
& 0.68
& 0.76
& 1.23 \\

\cline{2-5}
& \multicolumn{4}{l}{\textit{Questionnaire}} \\
\cline{2-5}
& Latency
& \textbf{6.50}
& 6.6
& \textbf{6.17} \\
& Responsiveness
& \textbf{5.83}
& 5.50
& \textbf{5.92} \\
& Ability to Correct
& \textbf{5.06}
& 5.28
& \textbf{5.56} \\
& Model Preference
& \textbf{6.33}
& 5.83
& \textbf{6.33} \\
\hline

\end{tabular}
\caption{User study results, including objective metrics and questionnaire scores. Questionnaire responses are grouped into Latency, Responsiveness, Ability to Correct, and Model Preference (7-point scale). We also report the mean number of points, mean number of undo actions, mean number of restarts, and mean interaction time (in seconds).}
\label{tab:merged_user_study}
\end{table*}

\paragraph{Summary}
Overall, the prospective user study demonstrates that both interactive methods substantially accelerated annotation compared with fully manual segmentation. While nnInteractive achieved superior performance for the MRI Uterus task, SLIP achieved comparable annotation efficiency for both liver lesion tasks while providing a more flexible interaction workflow through low-latency inference and reversible prompting. Together, the quantitative and user-reported findings suggest that practical annotation efficiency depends not only on segmentation accuracy, but also on interaction design, particularly support for low-latency iterative refinement and efficient recovery from intermediate interaction errors.

\paragraph{Ablations}

Because training a single SLIP model required several months, exhaustive retraining-based ablations were beyond the scope of this work. We therefore focused on the principal architectural component that can be isolated at inference time: the proposed patch feature contextualization mechanism.

Figure~\ref{fig:propagation_liver} illustrates the contribution of patch feature contextualization for a representative large anatomical structure (liver). Disabling patch feature contextualization substantially reduced segmentation quality during the earliest stages of refinement, whereas enabling it yielded an approximately 25\% improvement in DSC after the first interaction. This demonstrates that patch feature contextualization effectively propagates sparse user guidance across neighbouring patches, allowing large anatomical structures to be segmented efficiently from only a small number of user interactions. The benefit is greatest during the initial stages of refinement, where contextual information must be propagated over large spatial regions. As additional interactions are provided, the performance gap progressively narrows because subsequent user interactions directly correct the remaining local segmentation errors.

\section*{Conclusion}

We presented SLIP, an end-to-end framework for interactive 3D medical image segmentation that decouples image encoding from prompt-guided refinement through a lightweight patch memory bank. This architecture substantially reduces interaction latency while supporting reversible prompting, enabling efficient iterative refinement without recomputing image features.

An extensive simulated evaluation across 13 public datasets spanning diverse anatomical structures and imaging modalities demonstrated the robustness and generalizability of the proposed framework. SLIP achieved state-of-the-art interactive segmentation performance while substantially reducing interaction latency compared with existing methods. Although nnInteractive achieved stronger performance during the earliest refinement stages and on the MRI Uterus task, SLIP maintained responsiveness over prolonged interaction sequences and achieved higher final segmentation accuracy on the majority of the evaluated datasets.

The prospective user study complemented the large-scale simulated evaluation by assessing aspects of interactive segmentation that cannot be captured through simulated user interactions alone. Although nnInteractive achieved superior performance for the MRI Uterus task, participants consistently preferred low-latency interaction, iterative refinement, and reversible prompting, while SLIP eliminated the need to restart annotations and achieved higher overall user preference. Together, these findings demonstrate that practical annotation efficiency depends not only on segmentation accuracy, but also on the quality of the interaction workflow. In particular, low-latency interaction and reversible prompting enabled users to efficiently recover from intermediate interaction errors while preserving annotation progress during challenging cases.

Several limitations should be acknowledged. From a methodological perspective, point-based prompting remains inherently ambiguous during the earliest stages of interaction, particularly when multiple anatomically plausible structures surround the initial prompt, which can hinder optimization during training. Furthermore, the proposed framework relies on storing precomputed image patch embeddings throughout inference to achieve low-latency interaction. In practice, volumes up to approximately $512\times512\times512$ can be be processed within 10\,GB of GPU memory, encompassing the vast majority of routine clinical imaging volumes. Scaling to substantially larger volumes while preserving the same level of responsiveness may require more advanced memory management techniques, including GPU prefetching, host-memory caching, or asynchronous patch contextualization.

From an evaluation perspective, direct architectural comparison with existing foundation-style interactive segmentation methods remains challenging because identical training data and training pipelines are often unavailable. In particular, nnInteractive does not provide a publicly available training pipeline, and several datasets used during its original training are no longer publicly accessible. Consequently, the comparisons presented here should be interpreted as reflecting the performance of complete interactive segmentation systems rather than isolating architectural differences. Nevertheless, all methods were evaluated under identical inference conditions using the same datasets, interaction protocol, benchmarking hardware, and evaluation metrics, providing a fair and reproducible comparison of practical interactive performance.

Future work will investigate alternative interaction modalities, including scribble- and lasso-based prompting while preserving low-latency, reversible interaction paradigm introduced in this work. The decoupled architecture also provides a practical pathway for exploiting future advances in medical vision foundation models, enabling improved image encoders to be incorporated by retraining or fine-tuning only the lightweight prompt-guided refinement module rather than the entire interactive segmentation framework. More broadly, we believe that interactive segmentation systems should be evaluated not only by segmentation accuracy, but also by how effectively they support efficient and flexible human interaction through complementary large-scale simulated benchmarks and prospective user studies.

\section*{Acknowledgment}
We would like to acknowledge and thank Mathieu Haller, Cyriaque Zirimwabagabo, and Guinther Sa\"ibro for their invaluable assistance with data management and for the various fruitful discussions. We also thank Flavien Bridault and the IRCAD France R\&D software developers for valuable advice and support regarding software development. This work was supported by the French State through the Fonds national d'am\'enagement et de d\'eveloppement du territoire (FNADT), administered by the Prefecture of the Grand Est Region.

\section*{Declaration of generative AI and AI-assisted technologies in the manuscript preparation process}
During the preparation of this manuscript, the authors used ChatGPT (OpenAI) to assist with editing the manuscript to improve the clarity, conciseness, consistency, and grammatical correctness of the written English. ChatGPT was not used for any other aspect of manuscript preparation, including the generation or identification of references, data analysis, interpretation of results, or the formulation of scientific conclusions. All suggestions generated by ChatGPT were carefully reviewed and, where appropriate, edited before incorporation into the manuscript. The authors take full responsibility for the content of the published article.

\bibliographystyle{cas-model2-names}

\bibliography{cas-refs}



\end{document}


\section*{Supplementary Material}

\begin{table*}[h]
\centering
\tiny 
\setlength{\tabcolsep}{15pt} 
\begin{adjustbox}{max width=\textwidth}
\begin{tabular}{p{2.5cm} p{2.0cm} p{0.1cm} p{0.1cm} p{6.9cm}}
\hline
\textbf{Dataset} & \textbf{Structures} & \textbf{Img.} & \textbf{Mod.} & \textbf{Link} \\
\hline
Lymph Node CT Collection \cite{roth2018new}& Lymph nodes & 176 & CT & \url{https://www.cancerimagingarchive.net/collection/ct-lymph-nodes}\\
TCIA CervicalCancer \cite{mayr2023cervical} & Cervical tumor & 11 & MRI & \url{https://doi.org/10.7937/ERZ5-QZ59}\\
AeroPath \cite{stoverud2024aeropath} & Airway tree & 31 & CT & \url{https://github.com/raidionics/AeroPath}\\
Type-B Aortic Dissection \cite{mayer2024type} & Lumen, aorta & 44 & CT & \url{https://figshare.com/articles/dataset/Aortic_Dissection_Dataset_and_Segmentations/22269091}\\
COVID-19 CT Lung and Infection \cite{MP-COVID-19-SegBenchmark}& Lung, infection regions & 199 & CT & \url{http://medicalsegmentation.com/covid19}\\
CURVAS \cite{riera2025calibration}& Multiple abdominal organs & 94 & CT & \url{https://curvas.grand-challenge.org}\\
LCTSC \cite{yang2017data} & Lungs & 60 & CT & \url{https://doi.org/10.7937/K9/TCIA.2017.3R3FVZ08}\\
DAP-Atlas \cite{dap_atlas}& 142 anatomical structures & 533 & CT & \url{https://arxiv.org/abs/2307.13375}\\
TopCoW \cite{topcowchallenge}& Circle of Willis arteries & 250 & CT & \url{https://topcow23.grand-challenge.org}\\
AortaSeg24 \cite{imran2025multi}& Aorta & 100 & CT & \url{https://aortaseg24.grand-challenge.org}\\
COVID-19 Lung CT Scans \cite{mehrad_aria_mustafa_ghaderzadeh_farkhondeh_asadi_2021}& Lungs, COVID lesions & 349 & CT & \url{https://www.kaggle.com/dsv/1875670}\\
SAM-Med3D Collection \cite{wang2024sammed3dgeneralpurposesegmentationmodels} & Anatomical structures & 22k+ & CT/MRI. & \url{https://arxiv.org/abs/2310.15161}\\
MSWAL \cite{wu2025mswal} & Abdominal lesions & 484 & CT & \url{https://github.com/haochen-MBZUAI/MSWAL-}\\
DeepLesion \cite{yan2018deeplesion} & Universal lesions & 5000 & CT & \url{https://nihcc.app.box.com/v/DeepLesion}\\
CTSpine1K \cite{deng2021ctspine1k}& Spine & 1,005 & CT & \url{https://github.com/MIRACLE-Center/CTSpine1K}\\
AbdomenAtlas \cite{li2024abdomenatlas}& 25 abdominal organs & 5182 & CT & \url{https://github.com/MrGiovanni/AbdomenAtlas}\\
NSCLC-Radiomics \cite{aerts2015data}& Lung tumors & 415 & CT & \url{https://doi.org/10.7937/K9/TCIA.2015.PF0M9REI}\\
RibSeg \cite{yang2021ribseg}& Ribs & 370 & CT & \url{https://github.com/M3DV/RibSeg}\\
ToothFairy2 \cite{2024TMI} & Inferior alveolar canal, teeth & 480 & CT & \url{https://toothfairy.grand-challenge.org}\\
EMIDEC \cite{lalande2020emidec} & Myocardium, infarction scar & 100 & MRI & \url{https://emidec.com}\\
BrainMetShare \cite{grovik2020deep} & Brain metastases & 84 & MRI & \url{https://aimi.stanford.edu/brainmetshare}\\
Duke Liver \cite{duke_liver} & Liver & 310 & MRI & \url{https://zenodo.org/records/7774566}\\
OASIS \cite{marcus2007open}& Brain structures & 436 & MRI & \url{https://sites.wustl.edu/oasisbrains/home/oasis-1}\\
HNTS-MRG 2024 \cite{wahid2024overview} & Head and neck tumors & 248 & MRI & \url{https://hnts-mrg.grand-challenge.org}\\
HVSMR 2.0 \cite{pace2024hvsmr} & Whole heart structures & 120 & MRI & \url{https://doi.org/10.1038/s41597-024-03573-z}\\
TotalSegmentator MRI \cite{d2024totalsegmentator} & 80+ anatomical structures & 298 & MRI & \url{https://arxiv.org/abs/2405.19492}\\
SPIDER Lumbar Spine \cite{van2023spider}& Lumbar vertebrae and discs & 447 & MRI & \url{https://zenodo.org/records/10159290}\\
AutoPET \cite{gatidis2022whole}& Tumor lesions & 1,014 & PET/CT & \url{https://autopet-ii.grand-challenge.org/}\\
RESECT \cite{xiao2016resect}& Brain tumor & 69 & MRI+US & \url{https://osf.io/jv8bk}\\
CAMUS \cite{leclerc2019deep} & Heart & 1000 & US & \url{https://www.creatis.insa-lyon.fr/Challenge/camus}\\
TRUSTED \cite{ndzimbong2025trusted} & Kidney & 131 & US & \url{https://doi.org/10.1038/s41597-025-04892-5}\\
NIS3D \cite{zheng2023nis3d} & Cell nuclei & 6 & EM & \url{https://zenodo.org/records/11456029}\\
MUCIC HL60 \cite{svoboda2009generation} & Cell nuclei & 240 & VM & \url{https://cbia.fi.muni.cz/datasets/}\\
AxonEM \cite{wei2021axonem}& Axon instances & 18 & EM & \url{https://axonem.grand-challenge.org/}\\
NucMM \cite{lin2021nucmm} & Neuronal nuclei & 54 & EM & \url{https://nucmm.grand-challenge.org/}\\
SELMA3D \cite{SELMA3D-4} & Neural activity & 92 & Microscopy & \url{https://selma3d.grand-challenge.org/}\\
Urocell Endolysosomes \cite{ZerovnikMekuc2020}& Endolysosomes & 5 & Microscopy & \url{https://github.com/MancaZerovnikMekuc/UroCell}\\
Urocell Mitochondria \cite{ZerovnikMekuc2020} & Mitochondria & 5 & Microscopy & \url{https://github.com/MancaZerovnikMekuc/UroCell}\\
Cremi \cite{CREMI2016} & Neurons & 3 & Microscopy & \url{https://cremi.org/data/}\\
COVID-19-CT-Seg-Benchmark \cite{MP-COVID-19-SegBenchmark} & Lung, infection & 402 & CT & \url{https://github.com/JunMa11/COVID-19-CT-Seg-Benchmark}\\
LLD-MMRI \cite{LLD-MMRI} & Lung, infection & 3750 & MRI & \url{https://huggingface.co/datasets/wanglab/LLD-MMRI-MedSAM2}\\
CC Tumor Heterogeneity \cite{mayr2023cervical}& Cervix, tumor & 63 & MRI & \url{https://doi.org/10.7937/ERZ5-QZ59}\\
LIDC \cite{armato2015data} & Lung lesion & 1010 & CT & \url{https://www.cancerimagingarchive.net/collection/lidc-idri}\\
Atlas22 \cite{quinton2023tumor} & Stroke lesion & 524 & MRI & \url{https://atlas.grand-challenge.org/}\\
Prostate158 \cite{adams2022prostate158} & Prostate, tumors & 139 & MRI & \url{https://doi.org/10.5281/zenodo.6481141}\\
Flare 2023 \cite{ma2024automatic} & 13 abdominal organs & 2200 & CT & \url{https://codalab.lisn.upsaclay.fr/competitions/12239}\\
MAMA MIA \cite{garrucho2025large} & Breast lesions & 1506 & MRI & \url{https://www.synapse.org/Synapse:syn60868042/wiki/628716}\\
MMs \cite{martin2023deep}& LV/RV, myocardium & 300 & MRI & \url{https://www.ub.edu/mnms}\\
Leg-3D-US \cite{duque2024ultrasound} & Lower-limb leg & 42 & US & \url{https://www.cs.cit.tum.de/camp/publications/leg-3d-us-dataset/}\\
SegThy \cite{kronke2022tracked}& Neck & 32 & US & \url{https://www.cs.cit.tum.de/camp/publications/segthy-dataset/}\\
Thyroid Ultrasound Cine-clip \cite{stanford_cine}& Neck & 167 & US & \url{https://aimi.stanford.edu/datasets/thyroid-ultrasound-cine-clip}\\
Intra Operative liver lesions & Liver tumors & 555 & US & (private) \\
Fetus & Fetus & 64 & US & (private) \\
Kidney & Kidney & 300 & US & (private) \\
Liver & Liver & 1151 & US & (private) \\
\hline
\end{tabular}
\end{adjustbox}
\caption{Overview of the training datasets, including names, structures, number of images, modalities, and links }
\label{tab:medical_datasets}
\end{table*}

\begin{table*}[h]
\centering
\scriptsize
\begin{tabular}{p{4cm} p{5cm} p{2cm} p{2.5cm}}
\hline
\textbf{Dataset} & \textbf{Organs / Structures} & \textbf{\# Images} & \textbf{Modality} \\
\hline

SEGRAP \cite{luo2025segrap2023} & 45 Organs at Risk & 30 & CT \\

VTUS \cite{yang2025vivim} & Thyroid gland & 70 & Ultrasound \\

PENGWIN \cite{liu2023pelvic} & Pelvic bone fragments & 100 & CT \\

HANSEG \cite{podobnik2023han}& Head \& neck organs-at-risk & 42 & MRI \\

Colorectal livermets \cite{simpson2024preoperative} & Colorectal liver metastases & 171 & CT \\

Adrenal acc ki67 \cite{moawad2023voxel}& Adrenal glands / adrenocortical carcinoma & 53 & CT \\

Hcc tace liver \cite{moawad2023multimodality}& Liver (hepatocellular carcinoma) & 65 & CT \\

Hcc tace lesion \cite{moawad2023multimodality}& Hepatocellular carcinoma lesions & 65 & CT \\

RIDER Lung \cite{zhao2015coffee} & Lungs, pulmonary nodules & 59 & CT \\

LNQ \cite{dorent2024lnq}& Mediastinal lymph nodes & 513 & CT \\

TumSeg \cite{jensen20243d} & Subcutaneous tumors in mice & 452 & Micro-CT \\

UterUS \cite{bones2024automatic} & 3D uterine cavity & 141 & Ultrasound \\

TRUSTED \cite{ndzimbong2025trusted} & Kidneys & 48 & CT \\

\hline
\end{tabular}
\caption{Overview of the testing datasets, including names, structures, number of images, modalities}
\label{tab:extra_medical_datasets}
\end{table*}

\begin{figure*}[h]
\centering
\begin{Verbatim}[fontsize=\tiny, frame=single]
SLIP(
  ImageEncoder(
    Trunk: 3D-CNN(
      Stem: Conv3D(1, 64, k=3, s=1, p=1) + SyncBatchNorm + ReLU + MaxPool3D(k=3, s=2, p=1)
      Stage1 (x3 CNN blocks):
        CNNBlock(64 -> 64, k=3, s=1):
          Conv3D(k=3) + SyncBatchNorm + ReLU + Conv3D(k=3) + SyncBatchNorm  [+ residual]
      Stage2 (x4 CNN blocks):
        CNNBlock(64 -> 128, k=3, s=2, downsample):
          Conv3D(k=3,s=2) + SyncBatchNorm + ReLU + Conv3D(k=3) + SyncBatchNorm
          Downsample: Conv3D(1x1, s=2) + SyncBatchNorm  [residual proj]
        CNNBlock(128 -> 128, k=3, s=1)  [x3]:
          Conv3D(k=3) + SyncBatchNorm + ReLU + Conv3D(k=3) + SyncBatchNorm  [+ residual]
      Stage3 (x6 CNN blocks):
        CNNBlock(128 -> 256, k=3, s=2, downsample):
          Conv3D(k=3,s=2) + SyncBatchNorm + ReLU + Conv3D(k=3) + SyncBatchNorm
          Downsample: Conv3D(1x1, s=2) + SyncBatchNorm  [residual proj]
        CNNBlock(256 -> 256, k=3, s=1)  [x5]:
          Conv3D(k=3) + SyncBatchNorm + ReLU + Conv3D(k=3) + SyncBatchNorm  [+ residual]
      Stage4 (x3 CNN blocks):
        CNNBlock(256 -> 512, k=3, s=2, downsample):
          Conv3D(k=3,s=2) + SyncBatchNorm + ReLU + Conv3D(k=3) + SyncBatchNorm
          Downsample: Conv3D(1x1, s=2) + SyncBatchNorm  [residual proj]
        CNNBlock(512 -> 512, k=3, s=1)  [x2]:
          Conv3D(k=3) + SyncBatchNorm + ReLU + Conv3D(k=3) + SyncBatchNorm  [+ residual]
    )
    Neck: FPN(
      Takes multi-scale features from Stage4, Stage3, Stage2, Stage1 (channels 512, 256, 128, 64)
      Lateral convs (1x1 Conv3D, no norm/activation):
        L0: Conv3D(512 -> 256)   [from Stage4 output]
        L1: Conv3D(256 -> 256)   [from Stage3 output]
        L2: Conv3D(128 -> 256)   [from Stage2 output]
        L3: Conv3D(64  -> 256)   [from Stage1 output]
      Output: 4 feature maps, all with 256 channels, at strides {32, 16, 8, 4}
    )
  )
  PromptEncoder3D(
    Positional encoding: Random Fourier Features (3D)
    Visual Sampler: vis_sampler  (3D)
    Point embeddings: 2 x Embedding(1, 256)
    Mask downscaling: Conv3D(1->4, k=2,s=2) + LayerNorm3D + GELU
                     + Conv3D(4->16, k=2,s=2) + LayerNorm3D + GELU
                     + Conv3D(16->256, 1x1)
    No-mask embedding: Embedding(1, 256)
  )
  MemoryEncoder(
    MaskDownsampler: Conv3D(1->4, k=3,s=2) + LayerNorm3D + GELU
                   + Conv3D(4->16, k=3,s=2) + LayerNorm3D + GELU
                   + Conv3D(16->64, k=3,s=2) + LayerNorm3D + GELU
                   + Conv3D(64->256, 1x1)
    PixelFeatureProj: Conv3D(256 -> 256, 1x1)
    Fuser (x2 ConvNeXt blocks):
      DepthwiseConv3D(256, k=7, groups=256) + LayerNorm3D
      + Linear(256 -> 1024) + GELU + Linear(1024 -> 256)
    OutputProj: Conv3D(256 -> 64, 1x1)
  )
  PatchAttention (x4 layers):
    SelfAttention(256 -> 256) + LayerNorm + Dropout(0.1)
    CrossAttention(query=256, key/value=64 -> 256) + LayerNorm + Dropout(0.1)
    FeedForward: Linear(256 -> 2048) + ReLU + Linear(2048 -> 256) + LayerNorm + Dropout(0.1)
  MaskDecoder3D(
    conv_s0: Conv3D(256 -> 32, 1x1)   [high-res skip feature, stage 3]
    conv_s1: Conv3D(256 -> 64, 1x1)   [high-res skip feature, stage 4]
    TwoWayTransformer3D (x2 layers):
      SelfAttention(256 -> 256) + LayerNorm
      CrossAttention token->image (256 -> 128 -> 256) + LayerNorm
      MLP(256 -> 2048 -> 256) + ReLU + LayerNorm
      CrossAttention image->token (256 -> 128 -> 256) + LayerNorm
    Final CrossAttention token->image (256 -> 128 -> 256) + LayerNorm
    ClassificationToken: Embedding(1, 256)
    IoUToken: Embedding(1, 256)
    MaskTokens: Embedding(2, 256)
    OutputUpscaling:
      ConvTranspose3D(256 -> 64, k=2, s=2) + LayerNorm3D + GELU
      ConvTranspose3D(64 -> 32, k=2, s=2) + GELU
    MLPs (x2): Linear(256->256) + Linear(256->256) + Linear(256->32)
    IoUPredictionHead: Linear(256->256) + Linear(256->256) + Linear(256->2)
    ClassificationHead: Linear(256->256) + Linear(256->256) + Linear(256->1)
  )
)
Total parameters 75.0 M
\end{Verbatim}
\caption{Architecture of the SLIP model. CNN blocks follow a residual structure with two $3\times3\times3$ convolutions and SyncBatchNorm; downsampling blocks additionally project the residual path via a $1\times1\times1$ strided convolution. The FPN neck projects multi-scale trunk features to a common 256-channel dimension via $1\times1\times1$ lateral convolutions. ConvNeXt fuser blocks use depthwise convolutions followed by a channel-wise MLP. Attention modules (PatchAttention, TwoWayTransformer3D) are followed by LayerNorm and dropout where applicable.}
\label{fig:SLIP-arch}
\end{figure*}

\begin{table*}[h]
\centering
\label{tab:train_hparams}
\small
\begin{tabular}{p{3.2cm}|p{3.5cm}p{8.5cm}}
\toprule
\textbf{Phase} & \textbf{Hyperparameter} & \textbf{Value} \\
\midrule
\multirow{17}{*}{Phase 1}
 & GPUs & 4 GPUs \\
 & Batch size & 32 \\
 & Image size & $32 \times 192 \times 192$ (z, x, y) \\
 & Num epochs & 300  \\
 & Learning rate & $4\times10^{-5}$ \\
 & Optimizer & AdamW, $\beta=(0.9, 0.999)$ \\
 & LR scheduler & CosineAnnealingLR, $\eta_{\min}=\mathrm{lr}/20$ \\
 & Precision & bfloat16 \\
 & Gradient clipping & type: $\ell_2$, max: 0.1 \\
& Points per patch & 16 \\
\midrule
\multirow{16}{*}{Phase 2 }
 & GPUs  & 4 GPUs \\
 & One patch batch size & 32  \\
 & Propagation batch size & 8 \\
 & Image size & $32 \times 192 \times 192$ (z, x, y) \\
 & Num epochs / max iters per epoch & 500 (default) / 500 (default) \\
 & Learning rates & main$=8\times10^{-5}$, img. enc$=2\times10^{-5}$ \\
 & Optimizer & AdamW $\beta=(0.9, 0.999)$ \\
 & LR scheduler &CosineAnnealingLR, $\eta_{\min}=\mathrm{lr}/10$ \\
 & Dataset mixing (train / val) & [0.6, 0.4] / [0.5, 0.5] (prop, one patch) \\
 & Precision & bfloat16 \\
 & Gradient clipping & type: $\ell_2$, max: 0.1 \\
& Points per patch & 16 \\
& Corrected patches & 2 \\

\bottomrule

\end{tabular}
\caption{Training hyperparameters by training phase}

\end{table*}

\begin{table*}[t]
\centering
\label{tab:augs_gpu}
\small
\begin{tabular}{p{3.2cm}p{4.5cm}p{8.cm}}
\toprule
\textbf{Condition} & \textbf{Transform} & \textbf{Parameters} \\
\midrule
\multirow{12}{*}{All datasets (train)}
 & Resize (if oversized) & max\_size=(720,720,720) \\
 & Intensity clamping & range=[-1000, 1000] \\
 & Random resampling (spacing) & scale\_range=(0.8,1.25), anisotropic=True, $p=0.2$ \\
 & Random flip & axes=[1,2,3], $p=0.5$ \\
 & Random rotation & angle\_range=(-180,180), axes=['z'], $p=0.4$ \\
 & Random brightness (multiplicative) & multiplier\_range=(0.80,1.10), $p=0.15$ \\
 & Random contrast & contrast\_range=(0.65,1.5), $p=0.15$ \\
 & Random gamma correction & log\_gamma=(-0.5,0.5), $p=0.3$ \\
 & Random Gaussian noise & std=(0,0.5), $p=0.3$ \\
 & Random Gaussian blur & std=(0,2), $p=0.3$ \\
 & Z-score normalization & $p=1.0$ \\
\midrule
Non-US datasets (train) & Random elastic deformation & magnitude=2.0, control\_grid=16, $p=0.5$ \\
\midrule
\multirow{3}{*}{MRI datasets (train)}
 & Random MRI bias field & coefficients=0.5, $p=0.3$ \\
 & Random MRI ghosting artifact & num\_ghosts=4, intensity=(0.2,0.6), $p=0.15$ \\
 & Random MRI spike (k-space) artifact & num\_spikes=1, intensity=(0.1,0.4), $p=0.1$ \\
\midrule
\multirow{3}{*}{Validation / inference}
 & Resize (if oversized) & max\_size=(720,720,720) \\
 & Intensity clamping & range=[-1000, 1000] \\
 & Z-score normalization & $p=1.0$ \\
\bottomrule

\end{tabular}
\caption{Data augmentations used in the GPU pipeline during training}

\end{table*}

 
\begin{table*}[t]
\centering

\label{tab:questionnaire_uterus}
\renewcommand{\arraystretch}{1.25}
\setlength{\tabcolsep}{5pt}
\begin{tabularx}{\linewidth}{@{} c l X c c @{}}
\toprule
\textbf{\#} & \textbf{Type} & \textbf{Question} & \textbf{Scale 1} & \textbf{Scale 7} \\
\midrule
Q1 & Usability     & The GUI was easy to use.
    & \textit{Strongly disagree} & \textit{Strongly agree} \\
\midrule
Q2 & Latency       & How often were you frustrated by the model's waiting time?
    & \textit{Never} & \textit{Always} \\
Q3 & Latency       & To what extent does waiting time affect your workflow?
    & \textit{Not at all} & \textit{Completely disrupts} \\
Q4 & Latency  & How do you feel while waiting for the result?
    & \textit{Don't notice} & \textit{Very frustrated} \\
\midrule
Q5 & Responsiveness & During the first 10 clicks, the model was responsive.
    & \textit{Strongly disagree} & \textit{Strongly agree} \\
Q6 & Responsiveness & After the first 10 clicks, the model was still responsive.
    & \textit{Strongly disagree} & \textit{Strongly agree} \\
\midrule
Q7 & Ability to correct & After the first 10 clicks, the model produced a reasonable
                      interpretation, even if other segmentations were possible.
    & \textit{Strongly disagree} & \textit{Strongly agree} \\
Q8  & Ability to correct & How often could you fix incorrect segmentations with
                       positive/negative clicks?
    & \textit{Never} & \textit{Always} \\
Q9 & Ability to correct & When correcting a region, how often did clicks
                       affect other regions?
    & \textit{Never} & \textit{Always} \\
\midrule
Q10 & Overall satisfaction & I would consider using this system for annotation.
    & \textit{Strongly disagree} & \textit{Strongly agree} \\
\bottomrule

\end{tabularx}
\caption {Task-specific questionnaire.
         Items are rated on a 7-point Likert scale independently
         for \textbf{nnInteractive} and \textbf{SLIP}.}
\end{table*}
 
\begin{table*}[t]
\centering
\renewcommand{\arraystretch}{1.25}
\setlength{\tabcolsep}{5pt}
\begin{tabularx}{\linewidth}{@{} c >{\raggedright\arraybackslash}X c c @{}}
\toprule
\textbf{\#}  & \textbf{Question} & \textbf{Scale 1} & \textbf{Scale 7} \\
\midrule
Q1 & Low waiting time improves efficiency in interactive annotation tasks (waiting time = time between clicking and seeing the updated segmentation).
    & \textit{Strongly disagree} & \textit{Strongly agree} \\
Q2 & Low waiting time reduces frustration in interactive annotation tasks.
    & \textit{Strongly disagree} & \textit{Strongly agree} \\
Q3 & Low waiting time encourages users to spend more time refining a segmentation, instead of settling for a 'good enough' prediction. 
    & \textit{Strongly disagree} & \textit{Strongly agree} \\
Q4 & A model that can be iteratively refined to match the desired segmentation, even if it requires more interactions, is preferable to a model that performs well initially but offers limited refinement capability.
    & \textit{Strongly disagree} & \textit{Strongly agree} \\
Q5 & Having an undo feature is important.
    & \textit{Strongly disagree} & \textit{Strongly agree} \\
Q6 & Why is an undo feature important?
    & \multicolumn{2}{>{\centering\arraybackslash}p{0.34\linewidth}}{\textit{Multiple choice: correct mistakes easily; feel more confident using the system; experiment without risk; save time; reduce frustration; other}} \\
\bottomrule
\end{tabularx}
\caption{End-of-study questionnaire. Items Q1--Q5 are rated on a 7-point Likert scale. Q6 is a multiple-choice question.}
\label{tab:questionnaire}
\end{table*}

\clearpage

\bibliographystyle{cas-model2-names}

\bibliography{references}